\documentclass{article}

\usepackage[final]{neurips_2025}

\usepackage[utf8]{inputenc} 
\usepackage[T1]{fontenc}    
\usepackage{hyperref}       
\usepackage{url}            
\usepackage{booktabs}       
\usepackage{amsfonts, amsmath}       
\usepackage{nicefrac}       
\usepackage{microtype}      
\usepackage{xcolor}         
\newcommand{\colorword}[2]{%
    \textcolor{red!#1}{#2}%
}
\usepackage{graphicx}
\usepackage{listings}
\usepackage{subcaption}
\definecolor{myblue}{RGB}{108, 142, 191}   
\definecolor{mypurple}{RGB}{150, 115, 166} 

\DeclareMathOperator{\Corr}{Corr}

\title{Transferring Linear Features Across Language Models With Model Stitching}

\author{%
  Alan Chen\\
  Brown University\\
  \texttt{alan\_chen1@brown.edu} \\
  \And
    Jack Merullo \\
  Goodfire \\
  \texttt{jack@goodfire.ai} \\
   \AND
   Alessandro Stolfo \\
   ETH Z{\"u}rich \\
   \texttt{stolfoa@ethz.ch} \\
  \And
    Ellie Pavlick \\
   Brown University \\
   \texttt{ellie\_pavlick@brown.edu}
}

\begin{document}

\maketitle

\begin{abstract}
 In this work, we demonstrate that affine mappings between residual streams of language models is a cheap way to effectively transfer represented features between models. We apply this technique to transfer the \textit{weights} of Sparse Autoencoders (SAEs) between models of different sizes to compare their representations. We find that small and large models learn similar representation spaces, which motivates training expensive components like SAEs on a smaller model and transferring to a larger model at a FLOPs savings. In particular, using a small-to-large transferred SAE as initialization can lead to 50\% cheaper training runs when training SAEs on larger models. Next, we show that transferred probes and steering vectors can effectively recover ground truth performance. Finally, we dive deeper into feature-level transferability, finding that semantic and structural features transfer noticeably differently while specific classes of functional features have their roles faithfully mapped. Overall, our findings illustrate similarities and differences in the linear representation spaces of small and large models and demonstrate a method for improving the training efficiency of SAEs. 
\end{abstract}

\section{Introduction}
\label{sec:intro}

Large language models (LLMs) have continually proven to be surprisingly intricate next token predictors that become more complex yet predictably better with scale \citep{kaplan2021scaling}, leaving a desire to characterize and \textit{explain} the LLM's computations \citep{saphra2024mechanistic, olah2020zoom}. Despite increasing efforts to study the rich internal mechanisms and representations of LLMs, the full computations still remain opaque \citep{sharkey2025open, engels2024decomposing}. 

A popular perspective to view the internal computation of LLMs is through the lens of features, or functions of the input that serve a downstream purpose like predicting the next token or composing to form model circuits \citep{olah2020zoom, huben2024sparse}. The \textbf{Linear Representation Hypothesis} (LRH) posits that these features are represented as directions in a high dimensional space \citep{park2023linear, elhage2022toy}. However, the features had been empirically observed to appear in superposition (activating with interference), making interpretation of residual stream activations difficult. Sparse dictionary learning methods \citep{olshausen1996emergence, faruqui2015sparse, serre2006learning} such as Sparse Autoencoders (SAEs) have recently gained popularity for interpreting models through this lens by disentangling dense representations into sparsely activating feature sets \citep{huben2024sparse, bricken2023monosemanticity, gao2024scaling, kissane2024sae_finetuning}.
A second hypothesis, \textbf{Strong Model Universality} \citep{li2015convergent}, predicts that good models learn the same data representations.

While it is unlikely that the strongest versions of these hypotheses faithfully reflect how these models work (that is, every feature is represented as a direction and every feature should be represented universally across good models) \citep{engels2025not, csordas2024recurrent, wei2022emergent}, there is significant evidence to support weaker versions of both hypotheses: linear features have been used to probe and intervene on model representations \citep[i.a.]{zou2023universal, panickssery2023steering} and are the fundamental idea underlying SAEs. One form of evidence for universality is model stitching, which learns a transformation mapping the latent spaces of two models \citep{lenc2015understanding, bansal2021revisiting}. If the transformation is sufficiently simple, then it is claimed that the models encode information similarly. Other evidence includes recent work in ``model diffing'' \citep{lindsey2024crosscoders} using crosscoders, a variant of SAEs, and transferring SAEs between base and chat models \citep{kissane2024sae_finetuning} to study the effects of post-training.

Our work relies on the ansatz that if two models represent enough features as directions in similarly organized spaces, we should be able to transfer interpretable linear features between them. We propose the use of model stitches as the mappings for learned representations between language models. Our contributions (Figure \ref{fig:fig1}) can be summarized as follows:
\begin{enumerate}
    \item We learn stitches to transfer SAEs (\S \ref{sec:transferring_saes}) trained on one model to another, which are used to initialize SAE training runs on larger models with those trained on smaller ones (\S \ref{sec:stitching_saes}), allowing for an 50\% FLOPs savings. We also demonstrate that stitches can be used to transfer probes and steering vectors (\S \ref{sec:applications}).
    \item The stitches are affine mappings trained on within-family model pairs in Pythia \citep{biderman2023pythia}, GPT2 \citep{radford2019language}, and Gemma-2 \citep{team2024gemma}. We find downstream metrics are preserved by the mappings, supporting weak universality (\S \ref{sec:model_stitching}).
    \item We perform a fine-grained analysis on feature transfer and analyze how well semantic and structural features transfer using the stitch (\S \ref{sec:semantic_vs_structural}). We also find SAE features representing well-known universal functional features and note their transferability (\S \ref{sec:functional_features}).
\end{enumerate}
\section{Language Model Stitching}
\label{sec:model_stitching}
\begin{figure}
    \centering
    \includegraphics[width=1.0\linewidth]{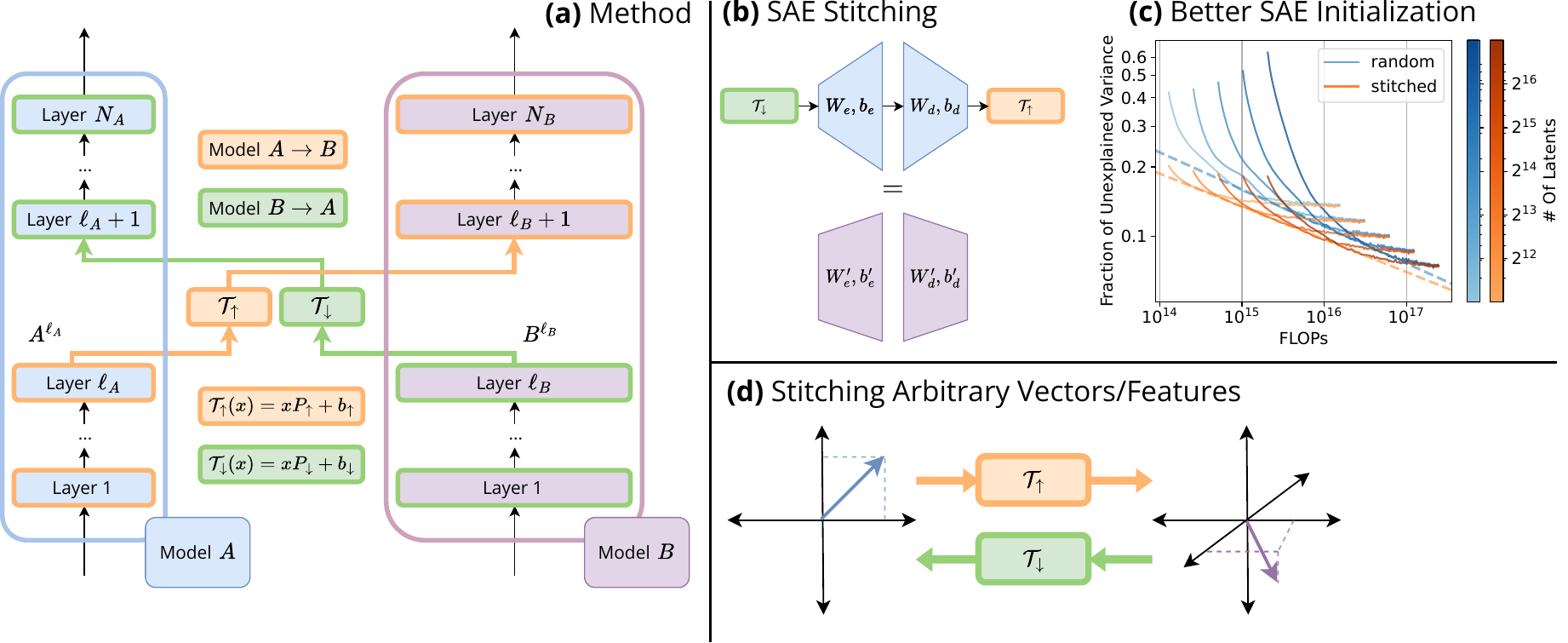}
    \caption{Overview of the main methodologies. \textbf{(a)} We train two affine mappings $\mathcal{T}_{\uparrow, \downarrow}$ concurrently to map between the residual streams of two language models $A$ and $B$. The mappings $\mathcal{T}_{\uparrow, \downarrow}$ are then used to transfer \textbf{(b)} the weights of entire SAEs from $A$ to $B$, which \textbf{(c)} give better initializations that save compute when training SAEs on $B$. The approximate ``scaling law'' for SAE training is shifted to the left when training from the transferred initialization, capturing the intuition that the transferred initialization saves the work of relearning shared features. More generally, the stitches can be used to transfer \textbf{(d)} arbitrary vectors (probes, steering vectors) between the residual stream spaces.}
    \label{fig:fig1}
\end{figure}
We would like to learn a mapping from the residual stream of language model $A$ to model $B$ at some layers, such that computation can start in the layers of $A$ and finish in $B$ (see Figure \ref{fig:fig1}a).
As notation, let $\textbf{LM}^{\ell}(t) \in \mathbb{R}^{d_\textbf{LM}}$ denote the representation in model $\textbf{LM}$ of input token(s) $t$ which has hidden state dimension $d_\textbf{LM}$ at layer $\ell$. Model stitching learns a mapping $\mathcal{T}$ (the \textbf{stitch}) between two model-layer pairs $(A, \ell_A)$ and $(B, \ell_B)$. Informally, we would like $\left[\mathcal{T} \circ A^{\ell_A}\right](t) \approx B^{\ell_B}(t)$ where $\circ$ is function composition. The existence of such a mapping $\mathcal{T}$ relies on at least weak universality, especially if $\mathcal{T}$ preserves the hypothesis classes of model $A$ and $B$. We are interested in the particular case when $A$ and $B$ are decoder-only language models from the same family (trained on the same data) but $d_B \geq d_A$. In this setup (visualized in Figure \ref{fig:fig1}a), we will consider the \textit{two} stitching mappings $\mathcal{T}_{\uparrow}$, which maps ``up'' from $A$ to $B$, and $\mathcal{T}_{\downarrow}$, which maps ``down'' from $B$ to $A$. Furthermore, we will assume that both $\mathcal{T}_\uparrow$ and $\mathcal{T}_\downarrow$ are affine transformations:
\begin{align}\label{eq:affine_Ts}
    \mathcal{T}_\uparrow: \mathbb{R}^{d_A} \to \mathbb{R}^{d_B}&, \quad h_A \mapsto h_A P_{\uparrow} + b_\uparrow, \\
    \mathcal{T}_\downarrow: \mathbb{R}^{d_B} \to \mathbb{R}^{d_A}&, \quad h_B \mapsto h_B P_{\downarrow} + b_\downarrow,
\end{align}
where $h_A$ and $h_B$ are activations from $A$ and $B$, $P_{\uparrow} \in \mathbb{R}^{d_A \times d_B}$, $b_{\uparrow} \in \mathbb{R}^{d_B}$, $P_{\downarrow} \in \mathbb{R}^{d_B \times d_A}$, and $b_\downarrow \in \mathbb{R}^{d_A}$. Ideally, we would like $\mathcal{T}_{\{\uparrow, \downarrow\}}$ to be faithful to the downstream objective (i.e., language modeling). In practice, training directly on the next-token prediction objective would involve backpropagating gradients over the back halves of $A$ and $B$, which is unnecessarily expensive. We find that the reconstruction mean squared error (MSE) is a sufficient training objective to align the models. Despite the dimensionality gap, we also still desire ``almost''-invertible transformations i.e.\ $\mathcal{T}_\uparrow \circ \mathcal{T}_\downarrow$ should be close to an identity and vice versa. Therefore, we also introduce a regularization penalty that encourages $\mathcal{T}_\uparrow$ and $\mathcal{T}_\downarrow$ to invert each other with relative strength $\alpha$, finding that the penalty improves fidelity (ablation in \S \ref{sec:inversion_ablation}) and has further motivation when we consider transferring SAEs in \S \ref{sec:stitching_saes} and \S \ref{sec:transferring_sae_appendix}. Formally, we train $\mathcal{T}_\uparrow$ and $\mathcal{T}_\downarrow$ concurrently via the loss function on a token $t$
\begin{align}\label{eq:stitching_loss_objective}
    \mathcal{L}(t) &= \text{MSE}\left(\left[\mathcal{T}_\uparrow \circ A^{\ell_A}\right](t), B^{\ell_B}(t)\right) +\text{MSE}\left(\left[\mathcal{T}_\downarrow \circ B^{\ell_B}\right](t), A^{\ell_A}(t)\right)\nonumber\\
    &+ \alpha\text{MSE}\left(\left[\mathcal{T}_\downarrow \circ \mathcal{T}_\uparrow \circ A^{\ell_A}\right](t), A^{\ell_A}(t)\right)+ \alpha \text{MSE}\left(\left[\mathcal{T}_\uparrow \circ \mathcal{T}_\downarrow \circ B^{\ell_B}\right](t), B^{\ell_B}(t)\right).
\end{align}

We always learn the mappings from a fixed residual stream layer of model $A$ to the layer in model $B$ that maximizes the average correlation from Singular Vector Canonical Correlation Analysis (SVCCA) across a small sample of model activations \citep{raghu2017svcca} (more details in \S \ref{sec:SVCCA}). Using this procedure, the model-layer pairs that we stitch between are shown in Table \ref{tab:model_layer_pairs}.
\begin{table}[]
\centering
\caption{Model-layer pairs that we stitch between and the associated downstream next token CE losses on OpenWebText. All activations are taken before the layer's computation. We primarily stitch from middle residual layers but for specific use cases (e.g., steering) we stitch between $\nicefrac{3}{4}$ of the depth. The losses are computed stitching from $A$ to $B$, $B$ to $A$, $A$ to $B$ back to $A$, and $B$ to $A$ back to $B$. We calculate relative losses to the model that comes last and color code with \textcolor{myblue}{blue} and \textcolor{mypurple}{purple} for model $A$ and model $B$ respectively.}
\label{tab:model_layer_pairs}
\begin{tabular}{@{}lcccccc@{}}
\toprule
\textbf{Model-Layer Pair}                                                                                                & $A$  & $B$  & $A \to B$ & $B \to A$ & $A \to B\to A$ & $B\to A \to B$ \\ \midrule
\begin{tabular}[c]{@{}l@{}}$A$: \textcolor{myblue}{pythia-70m-deduped.3}\\ $B$: \textcolor{mypurple}{pythia-160m-deduped.4}\end{tabular} & \textcolor{myblue}{3.60} & \textcolor{mypurple}{3.14} & \begin{tabular}[c]{@{}c@{}}3.99 \\(\textcolor{mypurple}{+27\%})\end{tabular}  & \begin{tabular}[c]{@{}c@{}}4.21 \\(\textcolor{myblue}{+17\%})\end{tabular} & \begin{tabular}[c]{@{}c@{}}3.62\\ (\textcolor{myblue}{+<1\%})\end{tabular} & \begin{tabular}[c]{@{}c@{}}3.29\\(\textcolor{mypurple}{+4.7\%}) \end{tabular}     \\ \midrule
\begin{tabular}[c]{@{}l@{}}$A$: \textcolor{myblue}{gpt2-small.6}\\ $B$: \textcolor{mypurple}{gpt2-medium.10}\end{tabular}                & \textcolor{myblue}{3.07}& \textcolor{mypurple}{2.77} & \begin{tabular}[c]{@{}c@{}}3.08\\ (\textcolor{mypurple}{+11\%})\end{tabular}       & \begin{tabular}[c]{@{}c@{}}3.35 \\ (\textcolor{myblue}{+9.1\%})\end{tabular}      & \begin{tabular}[c]{@{}c@{}} 3.07 \\ (\textcolor{myblue}{+<1\%})\end{tabular}          & \begin{tabular}[c]{@{}c@{}}2.81 \\(\textcolor{mypurple}{+1.4\%})\end{tabular}        \\\midrule
\begin{tabular}[c]{@{}l@{}}$A$: \textcolor{myblue}{gemma-2-2b.20}\\ $B$: \textcolor{mypurple}{gemma-2-9b.33}\end{tabular}                & \textcolor{myblue}{2.52} & \textcolor{mypurple}{2.36} & \begin{tabular}[c]{@{}c@{}} 3.28\\ (\textcolor{mypurple}{+39\%})\end{tabular}      & \begin{tabular}[c]{@{}c@{}}2.73\\ (\textcolor{myblue}{+8.3\%})\end{tabular}      & \begin{tabular}[c]{@{}c@{}}2.53\\(\textcolor{myblue}{+<1\%})\end{tabular}          & \begin{tabular}[c]{@{}c@{}}2.66\\ (\textcolor{mypurple}{+13\%})\end{tabular}           \\ \bottomrule
\end{tabular}
\end{table}

\paragraph{Downstream Performance.}
In Table \ref{tab:model_layer_pairs}, we compute the downstream fidelity of mapping from $A$ to $B$, mapping from $B$ to $A$, and mapping in both inverse directions. First, we observe that next-token prediction performance is close to ground truth, supporting a weak universality hypothesis that models trained on the same data share features despite differing scales. However, we also find that downstream performance is consistently bottlenecked by the worse (generally smaller) model $A$. This result suggests that the gap between how $A$ and $B$ use their latent spaces prevents linear stitching between $A$ and $B$ from recovering the full performance of $B$. Finally, the inverse operation can be nearly lossless despite the intrinsic dimension mismatch prohibiting perfect invertibility.

\section{Transferring SAEs}\label{sec:transferring_saes}
Sparse autoencoders (SAEs) have recently resurfaced in interpretability as a method of decomposing residual stream representations $x \in \mathbb{R}^d$ from a language model $\textbf{LM}$ at layer $\ell$ into sparse nonnegative feature activations $f(x; \theta) \in \mathbb{R}^{M}$, with $M \gg d$ and $\theta = (W_e, b_e, W_d, b_d)$ of the encoder matrix, encoder bias, decoder matrix, and decoder bias respectively that specify the affine transformations in the forward pass, given as
\begin{align}\label{eq:sae_forward_pass}
    f(x; \theta) &= \sigma(xW_e + b_e), \\
    \textbf{SAE}(x; \theta) &= f(x;\theta)W_d + b_d =\sum_{i=1}^{M} f_i(x;\theta) W_{d,i} + b_d.
\end{align}
The columns of the encoder matrix are the encoding/detection directions of the features. The rows of the decoder matrix are the decoding/representation directions of the features that the SAE decomposes into. Feature $i$ \textit{activates} on token $t$ if $f_i\left(\textbf{LM}^{\ell}(t); \theta\right) > 0$. SAEs are trained to reconstruct $x$ while having sparse activations via an objective function consisting of MSE between $x$ and $\textbf{SAE}(x;\theta)$ and, depending on the architecture, an $L_p$ regularization (usually with $p = 1$) on $f(x; \theta)$ with strength controlled via a parameter $\lambda > 0$, respectively. We use pretrained SAEs with TopK activation \citep{gao2024scaling} for Pythia\footnote{\url{https://github.com/EleutherAI/sparsify}} and JumpReLU SAEs for Gemma \citep{lieberum2024gemma} and train TopK SAEs from scratch on the GPT2 models using SAELens \citep{bloom2024saetrainingcodebase}.\footnote{\citet{gao2024scaling} releases pretrained TopK SAEs for GPT2. However, since these SAEs use a LayerNorm normalization scheme but our stitches are trained on unnormalized residual stream activations, the SAE transfer computation does not extend as easily.} 
\subsection{Efficient SAE Transfer}\label{sec:stitching_saes}
Equipped with the trained stitch transformations, we make a key observation that an SAE with parameters $\theta = (W_e, b_e, W_d, b_d)$ trained on layer $\ell_A$ in $A$ can be transferred to layer $\ell_B$ in $B$. This procedure is illustrated in Figure \ref{fig:fig1}b - specifically, for a latent $h_B = B^{\ell_B}(t)$ in model $B$, we (1) transfer down using $T_\downarrow$, (2) apply the original SAE parameterized by $\theta$, and (3) transfer back up using $T_\uparrow$. Because all the transformations are affine, this forward computation exactly specifies a new SAE on model $B$ parameterized by
\begin{equation}\label{eq:transferred_sae}
    \theta' = (W_e', b_e', W_d', b_d') = (P_\downarrow W_e, b_\downarrow W_e + b_e, W_dP_\uparrow, b_dP_\uparrow + b_\uparrow).
\end{equation}
This relationship is derived in detail in \S \ref{sec:transferring_sae_appendix}. The explicit formulas for $\theta'$ also provide geometric intuition for the parameters of $\mathcal{T}_{\{\uparrow, \downarrow\}}$. Indeed, $P_{\{\uparrow, \downarrow\}}$ can be viewed as directly transforming the linear feature spaces of $A$ and $B$ as they impact the feature vectors through $W_e'$ and $W_d'$. On the other hand, $b_{\{\uparrow, \downarrow\}}$ are responsible for adjusting the ``position'' of the feature decomposition as they only alter the biases $b_e'$ and $b_d'$.

We report evaluations of the transferred SAEs in \S \ref{sec:zero_shot_evaluations}, which are better than random but certainly worse than a fully trained SAE on $B$ (e.g., a Fraction Unexplained Variance [FUV] of 0.21 to 0.42 in Gemma 2B to 9B). The metrics serve as evidence of weak universality in the midst of a representational gap between $A$ and $B$. As a partial explanation, we note that the weight matrices of the transferred SAE are still rank $\leq d_A < d_B$ i.e., the SAE is only detecting and writing to features along a rank $d_A$ subspace embedded within the rank $d_B$ ambient embedding space of model $B$, contributing to their lackluster performance.

\paragraph{Training from a Stitch is More Efficient.}
 After obtaining a transferred SAE on $B$, we investigate how much continued training is required to match the performance of an SAE fully trained on model $B$. This question leads to a key application of the stitch and implicitly weak universality: \textbf{the transferred SAE can be used to initialize a new SAE in model $B$, ``saving'' the compute of having to relearn features that transfer well}. Concretely, consider the problem of training one SAE for two models of different sizes in the same family \citep{lieberum2024gemma}. The straightforward approach is to separately train a SAE from scratch on each model. However, we find that given a fully trained SAE on the small model, training a stitch and using the transferred small model SAE in the initialization scheme of the larger SAE allows for training of a comparable SAE to training from scratch but in a cheaper total FLoating point OPerations (FLOPs) budget.  We support this claim by training TopK SAEs on the Pythia model pair and repeat with GPT2 in \S\ref{sec:gpt2_sae_transfer}. We estimate FLOP counts based on the assumption that activations are cached prior to training both the stitch and the SAEs. All SAEs are trained using SAELens \citep{bloom2024saetrainingcodebase} and other training and FLOP estimation details are included in \S \ref{sec:sae_training_and_flop_estimation}.
 \begin{figure}
     \centering
     \includegraphics[width=0.9\linewidth]{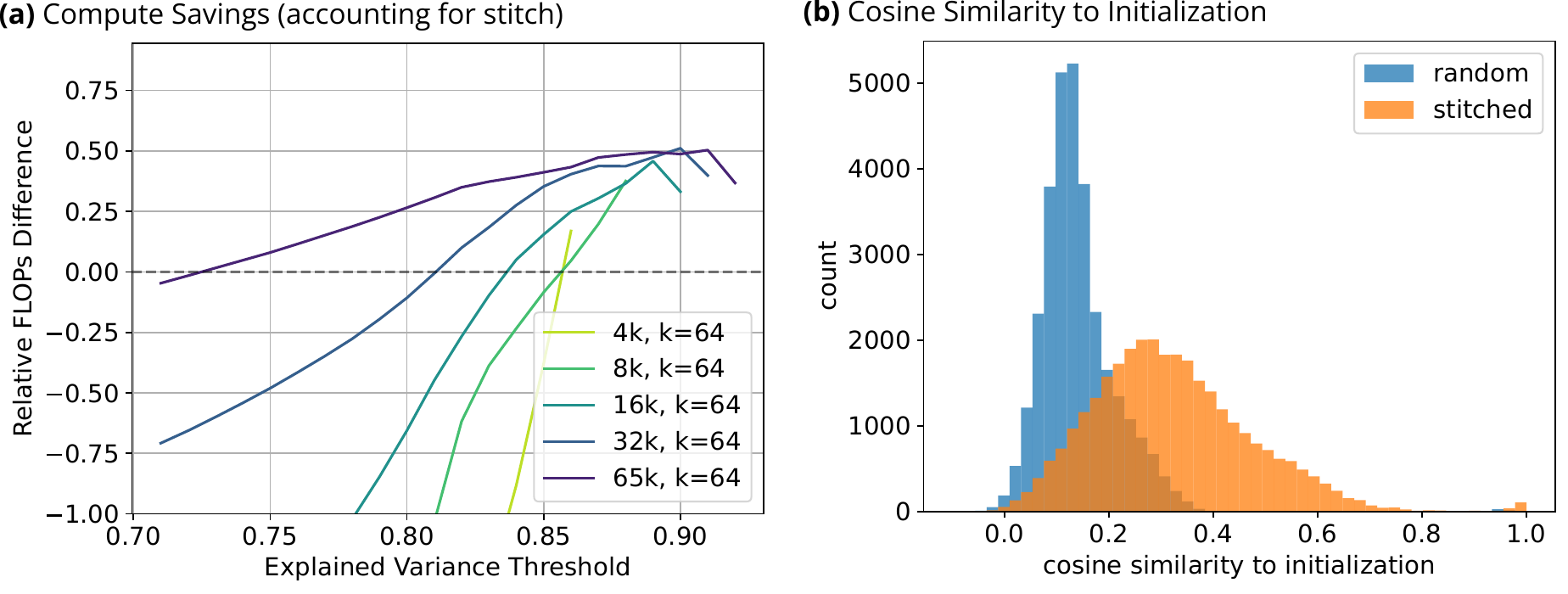}
    \caption{\textbf{(a)} In the Pythia model pair, transferred SAE initialization adjusted by the stitch FLOP count reaches explained variance thresholds in less FLOPs. For thresholds around $90\%$ explained variance, the moving average of explained variance of the SAE hits the threshold in around $30$-$50\%$ less FLOPs. \textbf{(b)} Features when trained from stitched initialization have higher cosine similarity to their initial state than random initialization in the $M = 32768$ runs. Dead features are removed from consideration for clarity around $1.0$.}
     \label{fig:other_scaling_visualizations}
 \end{figure}

Generally, we find that the cost of training the stitch on cached activations is much cheaper compared to training an SAE. Despite the cheap cost of the stitch, it allows us to stop the training of the SAE earlier when we use the transferred SAE as initialization. Figure \ref{fig:other_scaling_visualizations}a displays the relative FLOPs-to-hit-threshold savings of training from the transfer SAE initialization (accounting for the cost of training the stitch) vs. training from scratch initialization. More specifically, for the two initializations, we compare the number of FLOPs needed to get an SAE on Pythia-160m with explained variance above a particular threshold. At high explained variance values, we collect savings of between $30\%$-$50\%$ less FLOPs.

In Figure \ref{fig:fig1}c we repeat the experiment for various latent sizes to create an approximate scaling law plot inspired by \citet{gao2024scaling}. We emphasize our fit is only an \textit{approximation} meant for visualization as we estimate the law as a simple linear regression in log-log space instead of a proper scaling law (details and coefficients in \S \ref{sec:scaling_law_fitting}) - we ignore the irreducible loss term because we are comparing the two laws against each other and stably fitting the term requires larger latent sizes to saturate the curve. Nonetheless, when training from stitch initialization, the fitted law is noticeably shifted to the left compared to random initialization, indicating that the stitch initialization reaches levels of reconstruction loss relatively faster consistently across the given latent sizes. For small latent sizes, such a procedure is less worth the additional compute of training the stitch, as adding in the compute required to train the stitch would result in the procedure becoming more expensive than just training from scratch. To confirm our intuition about the initialization, we also compute cosine similarities of decoder vectors at the final checkpoint to their original direction at initialization. In Figure \ref{fig:other_scaling_visualizations}b, we observe that the features trained from stitch initialization rotate less in aggregate than training from random, indicating the initialization is indeed placing features closer to their final values. 

\section{Downstream Applications}\label{sec:applications}
Despite SAEs not transferring without additional training, the mappings $\mathcal{T}_\uparrow$ and $\mathcal{T}_\downarrow$ are fundamentally general transformations that linearly relate the residual streams of two models. In this section, we examine the application of the stitch to transferring probes and steering vectors, which target specific vectors in the residual stream (Figure \ref{fig:fig1}d). When applicable, following the intuition gained from Equation \ref{eq:transferred_sae}, we transfer feature vectors using just $P_{\uparrow, \downarrow}$ and do not include the bias. 
\begin{figure}
    \centering
    \includegraphics[width=1.0\linewidth]{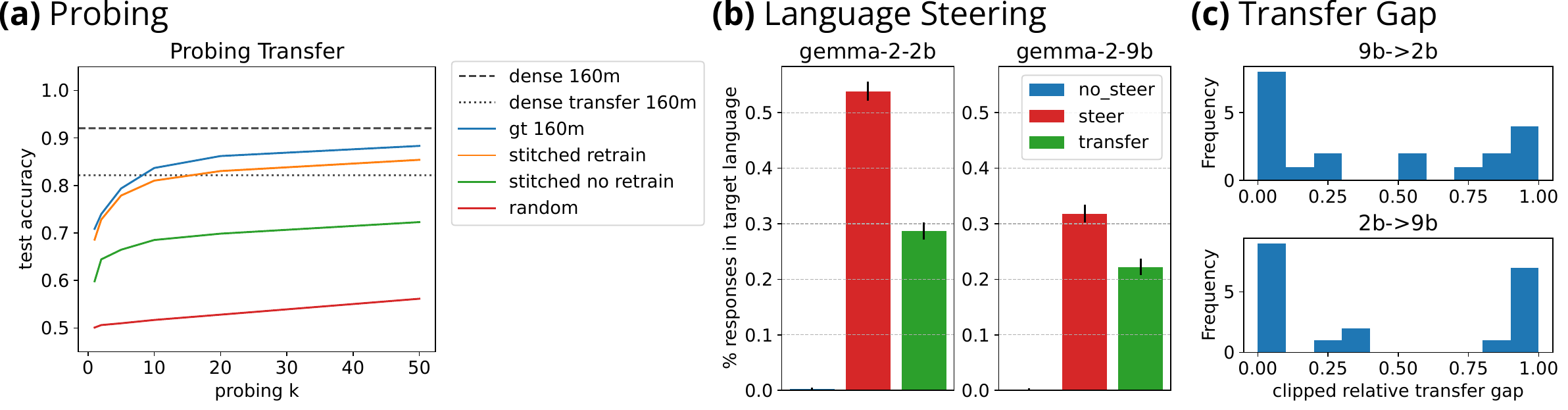}
    \caption{\textbf{(a)} Evaluations of transferred probes stitching from pythia-70m-deduped to pythia-160m-deduped averaged over $8$ binary classification datasets. If the probe is retrained (orange), we almost recover ground truth performance (blue) across all probing $k$s and most datasets. Even if probe is not retrained, in most datasets we are able to probe significantly better than random (green). When directly probing on the residual stream, we find that transferring a probe trained on 70m-deduped (dotted) reaches similar accuracy to a probe trained on 160m-deduped (dashed) without retraining. \textbf{(b)} Response language steering vectors are able to be transferred between gemma-2-2b.20 and gemma-2-9b.33. From left to right, we chart the \% of responses in the target language for no steering, ground truth steering, and steering using a transferred vector averaged over all languages $L$ in EuroParl and prompts. \textbf{(c)} The relative transfer gap distribution is bimodal with concentrations at $0$ and $1$, implying the transfer steering works well for some languages but poorly for others.}
    \label{fig:applications_fig}
\end{figure}
\paragraph{Probing.}
\label{sec:saebench}
To begin, we consider the sparse probing task \citep{gurnee2023finding, kantamneni2025sparse, karvonen2025saebench} - specifically, using a certain subset of $k$ SAE features as probing features for a binary concept. The $k$ features are selected from the features found by a SAE and are selected via a simple max mean activation difference procedure: feature activations are collected over training samples, averaged over non-padding tokens, and the class-wise difference in means is computed. The $k$ features with largest difference in means are used as the probing features. Formally, let $\mathcal{D}_+$ and $\mathcal{D}_-$ be the set of activations at a layer on positive and negative training examples respectively. Then, we compute the set $\mathcal{F}_k$ of indices of the $k$ features as
\begin{equation}\label{eq:topk_features}
    \mathcal{F}_k = \text{argTopK}_i\left(\mathbb{E}_{x_+ \sim \mathcal{D}_+}[f_i(x_+; \theta)] - \mathbb{E}_{x_- \sim \mathcal{D}_-}[f_i(x_-; \theta)]\right).
\end{equation}

In order to probe model $B$, we can transfer the SAE from $A$ to $B$ and use the activations from the same $k$ features in $\mathcal{F}_k$ to probe $B$. This procedure is equivalent to computing $\mathcal{T}_\downarrow(B^{\ell_B}(t))$, then applying a probe on $\mathcal{F}_k$'s activations on the stitched residual stream. Notably, the probe we use can either be completely reused as a pure evaluation of the stitch (i.e.\ the same probe that we trained when probing model $A$) or retrained to extract the full probing capability of the transferred features.

In Figure \ref{fig:applications_fig}a, we present the test accuracies for probes averaged across $8$ SAEBench datasets for $k = [1,2,5,10,20,50]$. The results for all datasets presented separately is in \S \ref{sec:full_probing_results}. We compare against baselines of selecting features from a ground truth SAE, random features, and dense representations in model $B$. On average, we are able to probe successfully using the transferred features \textit{without} retraining the probe - in particular, we do not collect activations from model $B$ outside of the data used to train the stitch. These results also hold with dense probes shown in the dashed and dotted lines: the dashed line is the skyline of training a probe on the dense representation of pythia-160m-deduped, whereas the dotted line is a probe trained on the dense representation of pythia-70m-deduped zero-shot applied to down-stitched residual streams from pythia-160m-deduped.

\paragraph{Steering/Instruction Following.}
In this section, we explore transferring steering vectors in a simple task: altering the response language from $\texttt{en}$ to a target language $L$ in EuroParl \citep{koehn2005europarl}. We compute a steering vector over the first $100$ paired examples in each \texttt{en}-$L$ dataset by unit-normalizing the difference in mean activation on $L$ tokens vs.\ English tokens \citep{panickssery2023steering}. To steer, we clamp the component in the feature's direction to $\bar{z}$, where we set $\bar{z}$ to the mean of $\langle \textbf{LM}^{\ell}(t), v\rangle$ over the tokens in $L$ (i.e., positive examples of the desired behavior). Explicitly, if $v$ is a steering vector and $\|v\|_2 = 1$, then we steer by modifying the hidden state at a token $t$ as
\begin{align}
    c &= \bar{z} - \langle \textbf{LM}^{\ell}(t), v \rangle,\\
    \textbf{LM}^{\ell}(t)' &= \textbf{LM}^{\ell}(t) + c v.
\end{align}
$v$ is transferred by computing $vP_{\{\uparrow,\downarrow\}}$, renormalizing, and recomputing $\bar{z}$ in the new model. We evaluate over a subset of $163$ prompts from the IFEval dataset with the instructions stripped and aggregate the proportion of responses in the target language \citep{zhou2023ifeval, stolfo2024improving}. Importantly, none of the prompts explicitly request the response to be in the target language. 

We use the Gemma model pair and present the results in Figure \ref{fig:applications_fig}b, where we plot the \% of responses in the target language depending on the type of intervention: no steering, steering with a vector learned from the current model, and steering with a transferred vector learned from the other model. Averaged over $L$, we find that the transferred steering vector identifies a direction that successfully steers the model toward responding in the target language without explicit prompt instruction to do so. We break down the overall accuracy into individual language pairs (\texttt{en}, $L$) and find the transferred steering vector works well for some languages but not for others by defining a \textit{clipped relative transfer gap} as the ratio of transfer steering performance to ground truth steering performance clipped to $[0,1]$ for visualization (Figure \ref{fig:applications_fig}c). We also note a positive correlation between language frequency and steering transfer effectiveness (\S \ref{sec:specific_language_results}). Although language steering appears to work somewhat well, we find that steering for general format instruction following has weaker results (\S \ref{sec:general_steering}).

\section{Feature Analysis}\label{sec:feature_analysis}
Finally, we can gain intuition into specifically what features transfer well under the stitch. To this end, we consider a correlation-based metric that approximates \textit{activation} and \textit{downstream effect} similarity \citet{bricken2023monosemanticity}. In particular, we compute the Pearson correlation between \textbf{attribution scores} induced by the logit weights and the model's next token predictions. 
Let $\text{Corr}(\cdot, \cdot)$ denote the Pearson correlation operator over data $x$ and let $v_{i}$ and $v_{i}'$ denote the logit weights given by unembedding the $i$th decoder vectors from $W_d$ and $W'_d$ respectively. The attribution correlation is given by 
\begin{equation}\label{eq:correlation_definition}
    \Corr_{t_\tau }\left(f_i\left(A^{\ell_A}(t_\tau);\theta\right)v_{i,t_{\tau+1, A}},f_i\left(B^{\ell_B}(t_\tau); \theta'\right) v'_{i, t_{\tau+1, B}}\right)
\end{equation}
where the correlation is computed over a set of tokens $t_\tau \in \mathcal{D}$ and $t_{\tau+1}$ is the next token. Intuitively, this score approximates the feature's importance in generating the next token prediction by taking the product of the activation and the downstream logit weight. We plot the histograms of attribution correlation scores against random SAEs in \S \ref{sec:attribution_correlation_histograms}. 

\subsection{Semantic vs. Structural features}\label{sec:semantic_vs_structural}
\begin{figure}
    \centering
    \includegraphics[width=0.9\linewidth]{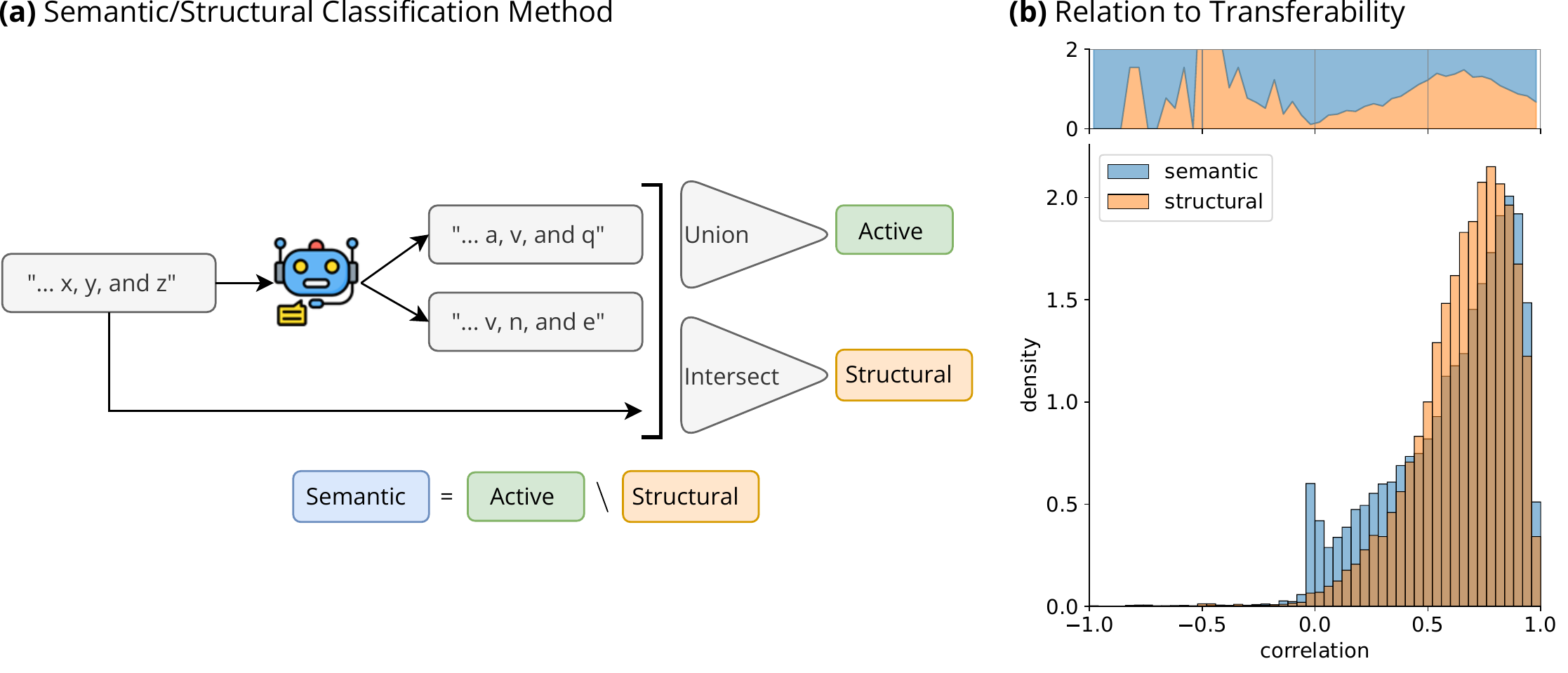}
    \caption{\textbf{(a)} An overview of the feature analysis pipeline for a simple example where $2$ augmentations are generated. Structural features activate on all prompts (intersection) whereas semantic features only activate on some but not all prompts. \textbf{(b)} The semantic/structural classification reveals a divergence in the attribution correlation transferability metric for non-dead features. We plot the densities of both categories separately and relative density above. Structural features transfer more consistently but semantic features are more polarized (dominate the upper and lower percentiles).}
    \label{fig:semantic_structural_experiment}
\end{figure}

The first question we can pose is whether different types of features are transferred relatively differently according to the metric in Equation \ref{eq:correlation_definition}. The simplest classes of features are semantic features vs.\ structural/syntactic features. We outline a cheap but general experiment to generate this separation, visualized in Figure \ref{fig:semantic_structural_experiment}a with example classifications in Table \ref{tab:cherrypicked_structural_vs_semantic}.
We first construct a synthetic dataset consisting of augmented versions of sentences drawn from a large text corpus. For each randomly drawn sentence, we use another LLM to generate $k$ augmented versions of the sentence with a prompt specifically instructing the LLM to ablate the semantic content of the sentence while maintaining the structure (\S \ref{sec:semantic_vs_structural_prompt}). We then feed the original sentence and all augmented versions ($1+k$ sentences in total) into the original language model and collect which SAE features activate on any token in the prompts. Features that activate consistently across the augmented versions of the same prompt are classified as \textbf{structural}, whereas all other features are classified as \textbf{semantic}. This classification is \textit{stitch-agnostic} - we classify the features purely based on activations from the original SAE. As a concrete example, suppose we began with the prompt ``I bought apples, bananas, and pears.'' We generate an augmented version of this prompt that preserves the structure of the sentence but ablates semantic content e.g.\ ``John fostered cats, dogs, and fish.'' Consider a feature that activates on the last comma of a list: this feature would activate in both prompts, resulting in classification as a structural feature. However, a feature that activates on food or animals would only activate on one of the prompts or the other, resulting in classification as a semantic feature.

We plot the histograms of structural vs. semantic features with respect to the attribution correlation in Figure \ref{fig:semantic_structural_experiment}b for the GPT2 stitch, removing all dead features from consideration. Our main observation is that structural features tend to consistently transfer better, but semantic features are more polarized: they generally transfer well or they do not. Directly, this result implies that the stitch strongly transfers a group of semantic features, mostly transfers structural features, and leaves behind some semantic features. This experiment also supports that new dimensions in larger models could be mostly allocated to developing increasingly specific semantic directions whereas the core language modeling spaces (which are more structural) remain similar.

\subsection{Functional Features}
\label{sec:functional_features}
\begin{figure}
    \centering
    \includegraphics[width=1.0\linewidth]{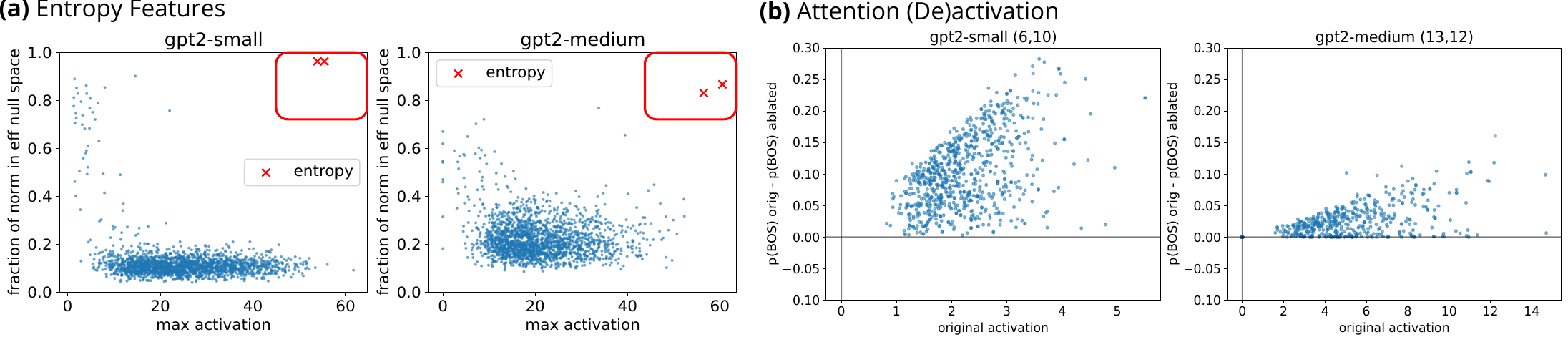}
    \caption{\textbf{(a)} Two entropy SAE features remain both large max activation and compose highly with the effective null space (bottom $2\%$ of singular values) before and after transfer. For clarity we take a randomly sampled subset of $2000$ features. \textbf{(b)} Attention pattern on BOS token on path patching experiment. After transfer, we are still able to find a head such that zero ablating the contribution of the feature has results in decreased attention on the BOS token. We only plot tokens in which the original feature activates in gpt2-small.}
    \label{fig:functional_features}
\end{figure}

Some of the structural features are related to known examples of universal features: in particular, entropy features and attention deactivation features \citep{gurnee2024universal}. We find that our stitch preserves the functional role of these features.
\paragraph{Entropy Features.} Entropy features are characterized by large norm and high composition with the effective null space (we define as the bottom $2\%$ of singular values) of the unembedding matrix \citep{stolfo2024confidence}. In gpt2-small, we find two features from an SAE that have these properties. Furthermore, we find that even after stitching, these properties still hold true (Figure \ref{fig:functional_features})a, showing that the stitch does indeed transfer the functional role of these features.
\paragraph{Attention Deactivation Features.} In order to identify attention deactivation neurons, we use the heuristic score and path patching techniques from \citet{gurnee2024universal}. In particular, to identify potential candidates for composition of a feature with a downstream attention head, we compute the heuristic score between all features and downstream attention heads. Then, to test a relationship between a feature and attention head, we path zero-ablate the feature's contribution to a target attention head's input at the current token over a small token set and measure the change in the attention pattern to the \texttt{<bos>} token. We identify an attention deactivation feature in the gpt2-small SAE and find that it retains its role as a deactivation feature for a downstream head post-transfer to gpt2-medium (Figure \ref{fig:functional_features}b). Attention sinks have also been noted to have large components to the effective null space \citep{cancedda-2024-spectral} and we find that is preserved as well (original: $0.27$, stitch: $0.77$). Somewhat mysteriously, we also observe the same feature acting as an attention \textit{activation} feature in gpt2-medium after transfer (Figure \ref{fig:attention_deactivation_flipped}).

\section{Related Work}
\paragraph{Language Model Representations.}
Language models have empirically been observed to learn vector representations of familiar human interpretable concepts like function application, truthfulness, factual knowledge, and refusal  \citep{todd2023function, gurnee2023language, zou2023universal, arditi2024refusal}. Linear directions can also be useful for model editing, steering, and concept erasure \citep{li2023inference, ilharco2022editing, panickssery2023steering, belrose2023leace}. 
\citep{huben2024sparse, bricken2023monosemanticity} reintroduced SAEs as a technique for decomposing residual streams which are hypothesized to geometrically superpose features \citep{elhage2022toy, arora2018linear, olah2020zoom}. SAEs have been somewhat successfully applied to various downstream tasks like steering, probing, circuit analysis, and discovering interesting features \citep{karvonen2025saebench, ameisen2025circuit, ferrando2024know, marks2024sparse}. Training SAEs has seen significant research effort, especially in architectural design and activation functions \citep{rajamanoharan2024jumping, gao2024scaling, bussmann2025learning}. Recently, however, limitations of SAEs have come into light and their use cases have become better understood \citep{paulo2025sparse, chanin2024absorption, kantamneni2025sparse, engels2024decomposing}. In particular, a consistent assumption across many aforementioned works is the linearity of features \citep{mikolov-etal-2013-linguistic, park2023linear, elhage2022toy}. However, recent work has demonstrated the existence of natural nonlinear features \citep{engels2025not, csordas2024recurrent}.

\paragraph{Universal Representations Across LMs.}
The intuition that well-generalizing models might all have similar representations has existed for quite some time \citep{li2015convergent, huh2024platonic}. Model universality has been studied from the perspective of model stitching \citep{lenc2015understanding, bansal2021revisiting, jha2025harnessing, csordas2025language}, representation similarity measures like SVCCA and CKA \citep{raghu2017svcca, kornblith2019similarity, barannikov2021representation}, SAEs and variants \citep{lindsey2024crosscoders, lan2024sparse}, and feature/neuron/weight level analysis \citep{gurnee2024universal, stolfo2024confidence}. It has also been leveraged in adversarial attack literature \citep{zou2023universal}. Recently, transfer of steering vectors between models of different sizes using linear mappings between the final residual streams has also been explored in binary steering tasks and safety applications \citep{lee2025shared, oozeer2025activation}. 
\section{Conclusion}
\paragraph{Findings.} We find that we can faithfully stitch between language models of different sizes of the same family, which can be leveraged to transfer SAEs, probes, and steering vectors. The SAEs can be trained to convergence faster than training from scratch, demonstrating an application that uses weak universality to benefit SAE training. Probes and steering vectors can also be transferred with no additional training in specific cases. Finally, we perform a case study on GPT2 and find differences between how the stitch transfers semantic and structural features and discover that specific functional features have their roles preserved.

\paragraph{Limitations and Future Work.} First, we only train the stitches on general internet text data between models in the same family with the same tokenizer. Natural follow ups are verifying the findings in a cross-family setting and finetuning the stitch to reconstruct unnaturally occurring special tokens or chat-templated data. The scaling laws are also theoretically incomplete because we lacked high-compute regime data to fit the irreducible loss. However, the transfer procedure is generalizable and other applications of the stitches to save compute or distill capabilities should be explored. For example, stitching could be used to transfer linear weight updates like LoRAs \citep{hu2022lora} which play nicely with the affine stitches and have already been established as similar to steering behavior \citep{engels2025awareness}.

The semantic/structural classification is also imperfect - we only gain one type of distinction which is also sometimes noisy because the ablations are generated by another language model and transferability is purely correlational. A closer analysis of sensitivity of the stitching and semantic/structural methodology to different layers could also produce new insights into feature distributions. Finally, the probing and steering experiments deserve to be expanded upon to examine robustness as our experiments are still limited to particular tasks and settings. 

\section*{Acknowledgments}
We would like to thank Michael Lepori, Jake Russin, and other members of the LUNAR Lab at Brown University for feedback at various stages of this project. We are also grateful to Neel Nanda for his valuable input during the early stages. AC would also like to thank Thomas Chang, Patrick Peng, and the CSCI2222 course at Brown for fruitful discussions about this work.  AS acknowledges the support of armasuisse Science and Technology through a CYD Doctoral Fellowship. This project was supported in part by a Young Faculty Award from the Defense Advanced Research Projects Agency Grant \#D24AP00261. Ellie Pavlick is a paid consultant for Google Deepmind. The content of this article does not necessarily reflect that of the US Government or of Google and no official endorsement of this work should be inferred.

\bibliography{refs}

\begin{thebibliography}{63}
\providecommand{\natexlab}[1]{#1}
\providecommand{\url}[1]{\texttt{#1}}
\expandafter\ifx\csname urlstyle\endcsname\relax
  \providecommand{\doi}[1]{doi: #1}\else
  \providecommand{\doi}{doi: \begingroup \urlstyle{rm}\Url}\fi

\bibitem[Ameisen et~al.(2025)Ameisen, Lindsey, Pearce, Gurnee, Turner, Chen, Citro, Abrahams, Carter, Hosmer, Marcus, Sklar, Templeton, Bricken, McDougall, Cunningham, Henighan, Jermyn, Jones, Persic, Qi, Ben~Thompson, Zimmerman, Rivoire, Conerly, Olah, and Batson]{ameisen2025circuit}
Emmanuel Ameisen, Jack Lindsey, Adam Pearce, Wes Gurnee, Nicholas~L. Turner, Brian Chen, Craig Citro, David Abrahams, Shan Carter, Basil Hosmer, Jonathan Marcus, Michael Sklar, Adly Templeton, Trenton Bricken, Callum McDougall, Hoagy Cunningham, Thomas Henighan, Adam Jermyn, Andy Jones, Andrew Persic, Zhenyi Qi, T.~Ben~Thompson, Sam Zimmerman, Kelley Rivoire, Thomas Conerly, Chris Olah, and Joshua Batson.
\newblock Circuit tracing: Revealing computational graphs in language models.
\newblock \emph{Transformer Circuits Thread}, 2025.
\newblock URL \url{https://transformer-circuits.pub/2025/attribution-graphs/methods.html}.

\bibitem[Arditi et~al.(2024)Arditi, Obeso, Syed, Paleka, Rimsky, Gurnee, and Nanda]{arditi2024refusal}
Andy Arditi, Oscar~Balcells Obeso, Aaquib Syed, Daniel Paleka, Nina Rimsky, Wes Gurnee, and Neel Nanda.
\newblock Refusal in language models is mediated by a single direction.
\newblock In \emph{The Thirty-eighth Annual Conference on Neural Information Processing Systems}, 2024.
\newblock URL \url{https://openreview.net/forum?id=pH3XAQME6c}.

\bibitem[Arora et~al.(2018)Arora, Li, Liang, Ma, and Risteski]{arora2018linear}
Sanjeev Arora, Yuanzhi Li, Yingyu Liang, Tengyu Ma, and Andrej Risteski.
\newblock Linear algebraic structure of word senses, with applications to polysemy.
\newblock \emph{Transactions of the Association for Computational Linguistics}, 6:\penalty0 483--495, 2018.

\bibitem[Bansal et~al.(2021)Bansal, Nakkiran, and Barak]{bansal2021revisiting}
Yamini Bansal, Preetum Nakkiran, and Boaz Barak.
\newblock Revisiting model stitching to compare neural representations.
\newblock \emph{Advances in neural information processing systems}, 34:\penalty0 225--236, 2021.

\bibitem[Barannikov et~al.(2021)Barannikov, Trofimov, Balabin, and Burnaev]{barannikov2021representation}
Serguei Barannikov, Ilya Trofimov, Nikita Balabin, and Evgeny Burnaev.
\newblock Representation topology divergence: A method for comparing neural network representations.
\newblock \emph{arXiv preprint arXiv:2201.00058}, 2021.

\bibitem[Belrose et~al.(2023)Belrose, Schneider-Joseph, Ravfogel, Cotterell, Raff, and Biderman]{belrose2023leace}
Nora Belrose, David Schneider-Joseph, Shauli Ravfogel, Ryan Cotterell, Edward Raff, and Stella Biderman.
\newblock {LEACE}: Perfect linear concept erasure in closed form.
\newblock In \emph{Thirty-seventh Conference on Neural Information Processing Systems}, 2023.
\newblock URL \url{https://openreview.net/forum?id=awIpKpwTwF}.

\bibitem[Biderman et~al.(2023)Biderman, Schoelkopf, Anthony, Bradley, O’Brien, Hallahan, Khan, Purohit, Prashanth, Raff, et~al.]{biderman2023pythia}
Stella Biderman, Hailey Schoelkopf, Quentin~Gregory Anthony, Herbie Bradley, Kyle O’Brien, Eric Hallahan, Mohammad~Aflah Khan, Shivanshu Purohit, USVSN~Sai Prashanth, Edward Raff, et~al.
\newblock Pythia: A suite for analyzing large language models across training and scaling.
\newblock In \emph{International Conference on Machine Learning}, pages 2397--2430. PMLR, 2023.

\bibitem[Bloom et~al.(2024)Bloom, Tigges, Duong, and Chanin]{bloom2024saetrainingcodebase}
Joseph Bloom, Curt Tigges, Anthony Duong, and David Chanin.
\newblock Saelens.
\newblock \url{https://github.com/jbloomAus/SAELens}, 2024.

\bibitem[Bricken et~al.(2023)Bricken, Templeton, Batson, Chen, Jermyn, Conerly, Turner, Anil, Denison, Askell, Lasenby, Wu, Kravec, Schiefer, Maxwell, Joseph, Hatfield-Dodds, Tamkin, Nguyen, McLean, Burke, Hume, Carter, Henighan, and Olah]{bricken2023monosemanticity}
Trenton Bricken, Adly Templeton, Joshua Batson, Brian Chen, Adam Jermyn, Tom Conerly, Nick Turner, Cem Anil, Carson Denison, Amanda Askell, Robert Lasenby, Yifan Wu, Shauna Kravec, Nicholas Schiefer, Tim Maxwell, Nicholas Joseph, Zac Hatfield-Dodds, Alex Tamkin, Karina Nguyen, Brayden McLean, Josiah~E Burke, Tristan Hume, Shan Carter, Tom Henighan, and Christopher Olah.
\newblock Towards monosemanticity: Decomposing language models with dictionary learning.
\newblock \emph{Transformer Circuits Thread}, 2023.
\newblock https://transformer-circuits.pub/2023/monosemantic-features/index.html.

\bibitem[Bussmann et~al.(2025)Bussmann, Nabeshima, Karvonen, and Nanda]{bussmann2025learning}
Bart Bussmann, Noa Nabeshima, Adam Karvonen, and Neel Nanda.
\newblock Learning multi-level features with matryoshka sparse autoencoders.
\newblock \emph{arXiv preprint arXiv:2503.17547}, 2025.

\bibitem[Cancedda(2024)]{cancedda-2024-spectral}
Nicola Cancedda.
\newblock Spectral filters, dark signals, and attention sinks.
\newblock In Lun-Wei Ku, Andre Martins, and Vivek Srikumar, editors, \emph{Proceedings of the 62nd Annual Meeting of the Association for Computational Linguistics (Volume 1: Long Papers)}, pages 4792--4808, Bangkok, Thailand, August 2024. Association for Computational Linguistics.
\newblock \doi{10.18653/v1/2024.acl-long.263}.
\newblock URL \url{https://aclanthology.org/2024.acl-long.263/}.

\bibitem[Chanin et~al.(2025)Chanin, Wilken-Smith, Dulka, Bhatnagar, and Bloom]{chanin2024absorption}
David Chanin, James Wilken-Smith, Tom{\'a}{\v{s}} Dulka, Hardik Bhatnagar, and Joseph~Isaac Bloom.
\newblock A is for absorption: Studying feature splitting and absorption in sparse autoencoders, 2025.
\newblock URL \url{https://openreview.net/forum?id=LC2KxRwC3n}.

\bibitem[Csord{\'a}s et~al.(2024)Csord{\'a}s, Potts, Manning, and Geiger]{csordas2024recurrent}
R{\'o}bert Csord{\'a}s, Christopher Potts, Christopher~D Manning, and Atticus Geiger.
\newblock Recurrent neural networks learn to store and generate sequences using non-linear representations.
\newblock In \emph{The 7th BlackboxNLP Workshop}, 2024.
\newblock URL \url{https://openreview.net/forum?id=NUQeYgg8x4}.

\bibitem[Csord{\'a}s et~al.(2025)Csord{\'a}s, Manning, and Potts]{csordas2025language}
R{\'o}bert Csord{\'a}s, Christopher~D Manning, and Christopher Potts.
\newblock Do language models use their depth efficiently?
\newblock \emph{arXiv preprint arXiv:2505.13898}, 2025.

\bibitem[Elhage et~al.(2022)Elhage, Hume, Olsson, Schiefer, Henighan, Kravec, Hatfield-Dodds, Lasenby, Drain, Chen, et~al.]{elhage2022toy}
Nelson Elhage, Tristan Hume, Catherine Olsson, Nicholas Schiefer, Tom Henighan, Shauna Kravec, Zac Hatfield-Dodds, Robert Lasenby, Dawn Drain, Carol Chen, et~al.
\newblock Toy models of superposition.
\newblock \emph{arXiv preprint arXiv:2209.10652}, 2022.

\bibitem[Engels et~al.(2025{\natexlab{a}})Engels, Nanda, and Rajamanoharan]{engels2025awareness}
Josh Engels, Neel Nanda, and Senthooran Rajamanoharan.
\newblock Interim research report: Mechanisms of awareness.
\newblock Alignment Forum, 2025{\natexlab{a}}.
\newblock URL \url{https://www.alignmentforum.org/posts/m8WKfNxp9eDLRkCk9/interim-research-report-mechanisms-of-awareness}.

\bibitem[Engels et~al.(2024)Engels, Riggs, and Tegmark]{engels2024decomposing}
Joshua Engels, Logan Riggs, and Max Tegmark.
\newblock Decomposing the dark matter of sparse autoencoders.
\newblock \emph{arXiv preprint arXiv:2410.14670}, 2024.

\bibitem[Engels et~al.(2025{\natexlab{b}})Engels, Michaud, Liao, Gurnee, and Tegmark]{engels2025not}
Joshua Engels, Eric~J Michaud, Isaac Liao, Wes Gurnee, and Max Tegmark.
\newblock Not all language model features are one-dimensionally linear.
\newblock In \emph{The Thirteenth International Conference on Learning Representations}, 2025{\natexlab{b}}.
\newblock URL \url{https://openreview.net/forum?id=d63a4AM4hb}.

\bibitem[Faruqui et~al.(2015)Faruqui, Tsvetkov, Yogatama, Dyer, and Smith]{faruqui2015sparse}
Manaal Faruqui, Yulia Tsvetkov, Dani Yogatama, Chris Dyer, and Noah~A Smith.
\newblock Sparse overcomplete word vector representations.
\newblock In \emph{Proceedings of the 53rd Annual Meeting of the Association for Computational Linguistics and the 7th International Joint Conference on Natural Language Processing (Volume 1: Long Papers)}, pages 1491--1500, 2015.

\bibitem[Ferrando et~al.(2025)Ferrando, Obeso, Rajamanoharan, and Nanda]{ferrando2024know}
Javier Ferrando, Oscar~Balcells Obeso, Senthooran Rajamanoharan, and Neel Nanda.
\newblock Do i know this entity? knowledge awareness and hallucinations in language models.
\newblock In \emph{The Thirteenth International Conference on Learning Representations}, 2025.
\newblock URL \url{https://openreview.net/forum?id=WCRQFlji2q}.

\bibitem[Gao et~al.(2025)Gao, la~Tour, Tillman, Goh, Troll, Radford, Sutskever, Leike, and Wu]{gao2024scaling}
Leo Gao, Tom~Dupre la~Tour, Henk Tillman, Gabriel Goh, Rajan Troll, Alec Radford, Ilya Sutskever, Jan Leike, and Jeffrey Wu.
\newblock Scaling and evaluating sparse autoencoders.
\newblock In \emph{The Thirteenth International Conference on Learning Representations}, 2025.
\newblock URL \url{https://openreview.net/forum?id=tcsZt9ZNKD}.

\bibitem[Gurnee and Tegmark(2024)]{gurnee2023language}
Wes Gurnee and Max Tegmark.
\newblock Language models represent space and time.
\newblock In \emph{The Twelfth International Conference on Learning Representations}, 2024.
\newblock URL \url{https://openreview.net/forum?id=jE8xbmvFin}.

\bibitem[Gurnee et~al.(2023)Gurnee, Nanda, Pauly, Harvey, Troitskii, and Bertsimas]{gurnee2023finding}
Wes Gurnee, Neel Nanda, Matthew Pauly, Katherine Harvey, Dmitrii Troitskii, and Dimitris Bertsimas.
\newblock Finding neurons in a haystack: Case studies with sparse probing.
\newblock \emph{Transactions on Machine Learning Research}, 2023.
\newblock ISSN 2835-8856.
\newblock URL \url{https://openreview.net/forum?id=JYs1R9IMJr}.

\bibitem[Gurnee et~al.(2024)Gurnee, Horsley, Guo, Kheirkhah, Sun, Hathaway, Nanda, and Bertsimas]{gurnee2024universal}
Wes Gurnee, Theo Horsley, Zifan~Carl Guo, Tara~Rezaei Kheirkhah, Qinyi Sun, Will Hathaway, Neel Nanda, and Dimitris Bertsimas.
\newblock Universal neurons in {GPT}2 language models.
\newblock \emph{Transactions on Machine Learning Research}, 2024.
\newblock ISSN 2835-8856.
\newblock URL \url{https://openreview.net/forum?id=ZeI104QZ8I}.

\bibitem[Hu et~al.(2022)Hu, yelong shen, Wallis, Allen-Zhu, Li, Wang, Wang, and Chen]{hu2022lora}
Edward~J Hu, yelong shen, Phillip Wallis, Zeyuan Allen-Zhu, Yuanzhi Li, Shean Wang, Lu~Wang, and Weizhu Chen.
\newblock Lo{RA}: Low-rank adaptation of large language models.
\newblock In \emph{International Conference on Learning Representations}, 2022.
\newblock URL \url{https://openreview.net/forum?id=nZeVKeeFYf9}.

\bibitem[Huben et~al.(2024)Huben, Cunningham, Smith, Ewart, and Sharkey]{huben2024sparse}
Robert Huben, Hoagy Cunningham, Logan~Riggs Smith, Aidan Ewart, and Lee Sharkey.
\newblock Sparse autoencoders find highly interpretable features in language models.
\newblock In \emph{The Twelfth International Conference on Learning Representations}, 2024.
\newblock URL \url{https://openreview.net/forum?id=F76bwRSLeK}.

\bibitem[Huh et~al.(2024)Huh, Cheung, Wang, and Isola]{huh2024platonic}
Minyoung Huh, Brian Cheung, Tongzhou Wang, and Phillip Isola.
\newblock The platonic representation hypothesis.
\newblock \emph{arXiv preprint arXiv:2405.07987}, 2024.

\bibitem[Ilharco et~al.(2023)Ilharco, Ribeiro, Wortsman, Schmidt, Hajishirzi, and Farhadi]{ilharco2022editing}
Gabriel Ilharco, Marco~Tulio Ribeiro, Mitchell Wortsman, Ludwig Schmidt, Hannaneh Hajishirzi, and Ali Farhadi.
\newblock Editing models with task arithmetic.
\newblock In \emph{The Eleventh International Conference on Learning Representations}, 2023.
\newblock URL \url{https://openreview.net/forum?id=6t0Kwf8-jrj}.

\bibitem[Jha et~al.(2025)Jha, Zhang, Shmatikov, and Morris]{jha2025harnessing}
Rishi Jha, Collin Zhang, Vitaly Shmatikov, and John~X Morris.
\newblock Harnessing the universal geometry of embeddings.
\newblock \emph{arXiv preprint arXiv:2505.12540}, 2025.

\bibitem[Kantamneni et~al.(2025)Kantamneni, Engels, Rajamanoharan, Tegmark, and Nanda]{kantamneni2025sparse}
Subhash Kantamneni, Joshua Engels, Senthooran Rajamanoharan, Max Tegmark, and Neel Nanda.
\newblock Are sparse autoencoders useful? a case study in sparse probing.
\newblock \emph{arXiv preprint arXiv:2502.16681}, 2025.

\bibitem[Kaplan et~al.(2020)Kaplan, McCandlish, Henighan, Brown, Chess, Child, Gray, Radford, Wu, and Amodei]{kaplan2021scaling}
Jared Kaplan, Sam McCandlish, Tom Henighan, Tom~B Brown, Benjamin Chess, Rewon Child, Scott Gray, Alec Radford, Jeffrey Wu, and Dario Amodei.
\newblock Scaling laws for neural language models.
\newblock \emph{arXiv preprint arXiv:2001.08361}, 2020.

\bibitem[Karvonen et~al.(2025)Karvonen, Rager, Lin, Tigges, Bloom, Chanin, Lau, Farrell, McDougall, Ayonrinde, Wearden, Conmy, Marks, and Nanda]{karvonen2025saebench}
Adam Karvonen, Can Rager, Johnny Lin, Curt Tigges, Joseph Bloom, David Chanin, Yeu-Tong Lau, Eoin Farrell, Callum McDougall, Kola Ayonrinde, Matthew Wearden, Arthur Conmy, Samuel Marks, and Neel Nanda.
\newblock Saebench: A comprehensive benchmark for sparse autoencoders in language model interpretability, 2025.
\newblock URL \url{https://arxiv.org/abs/2503.09532}.

\bibitem[Kissane et~al.(2024)Kissane, Krzyzanowski, Conmy, and Nanda]{kissane2024sae_finetuning}
Connor Kissane, Robert Krzyzanowski, Arthur Conmy, and Neel Nanda.
\newblock Saes (usually) transfer between base and chat models.
\newblock Alignment Forum, 2024.
\newblock URL \url{https://www.alignmentforum.org/posts/fmwk6qxrpW8d4jvbd/saes-usually-transfer-between-base-and-chat-models}.

\bibitem[Koehn(2005)]{koehn2005europarl}
Philipp Koehn.
\newblock {E}uroparl: A parallel corpus for statistical machine translation.
\newblock In \emph{Proceedings of Machine Translation Summit X: Papers}, pages 79--86, Phuket, Thailand, September 13-15 2005.
\newblock URL \url{https://aclanthology.org/2005.mtsummit-papers.11}.

\bibitem[Kornblith et~al.(2019)Kornblith, Norouzi, Lee, and Hinton]{kornblith2019similarity}
Simon Kornblith, Mohammad Norouzi, Honglak Lee, and Geoffrey Hinton.
\newblock Similarity of neural network representations revisited.
\newblock In \emph{International conference on machine learning}, pages 3519--3529. PMLR, 2019.

\bibitem[Lan et~al.(2024)Lan, Torr, Meek, Khakzar, Krueger, and Barez]{lan2024sparse}
Michael Lan, Philip Torr, Austin Meek, Ashkan Khakzar, David Krueger, and Fazl Barez.
\newblock Sparse autoencoders reveal universal feature spaces across large language models.
\newblock \emph{arXiv preprint arXiv:2410.06981}, 2024.

\bibitem[Lee et~al.(2025)Lee, Weber, Vi{\'e}gas, and Wattenberg]{lee2025shared}
Andrew Lee, Melanie Weber, Fernanda Vi{\'e}gas, and Martin Wattenberg.
\newblock Shared global and local geometry of language model embeddings.
\newblock \emph{arXiv preprint arXiv:2503.21073}, 2025.

\bibitem[Lenc and Vedaldi(2015)]{lenc2015understanding}
Karel Lenc and Andrea Vedaldi.
\newblock Understanding image representations by measuring their equivariance and equivalence.
\newblock In \emph{Proceedings of the IEEE conference on computer vision and pattern recognition}, pages 991--999, 2015.

\bibitem[Li et~al.(2023)Li, Patel, Vi{\'e}gas, Pfister, and Wattenberg]{li2023inference}
Kenneth Li, Oam Patel, Fernanda Vi{\'e}gas, Hanspeter Pfister, and Martin Wattenberg.
\newblock Inference-time intervention: Eliciting truthful answers from a language model.
\newblock In \emph{Thirty-seventh Conference on Neural Information Processing Systems}, 2023.
\newblock URL \url{https://openreview.net/forum?id=aLLuYpn83y}.

\bibitem[Li et~al.(2015)Li, Yosinski, Clune, Lipson, and Hopcroft]{li2015convergent}
Yixuan Li, Jason Yosinski, Jeff Clune, Hod Lipson, and John Hopcroft.
\newblock Convergent learning: Do different neural networks learn the same representations?
\newblock In Dmitry Storcheus, Afshin Rostamizadeh, and Sanjiv Kumar, editors, \emph{Proceedings of the 1st International Workshop on Feature Extraction: Modern Questions and Challenges at NIPS 2015}, volume~44 of \emph{Proceedings of Machine Learning Research}, pages 196--212, Montreal, Canada, 11 Dec 2015. PMLR.
\newblock URL \url{https://proceedings.mlr.press/v44/li15convergent.html}.

\bibitem[Lieberum et~al.(2024)Lieberum, Rajamanoharan, Conmy, Smith, Sonnerat, Varma, Kram{\'a}r, Dragan, Shah, and Nanda]{lieberum2024gemma}
Tom Lieberum, Senthooran Rajamanoharan, Arthur Conmy, Lewis Smith, Nicolas Sonnerat, Vikrant Varma, J{\'a}nos Kram{\'a}r, Anca Dragan, Rohin Shah, and Neel Nanda.
\newblock Gemma scope: Open sparse autoencoders everywhere all at once on gemma 2.
\newblock \emph{arXiv preprint arXiv:2408.05147}, 2024.

\bibitem[Lindsey et~al.(2024)Lindsey, Templeton, Marcus, Conerly, Baston, and Olah]{lindsey2024crosscoders}
Jack Lindsey, Adly Templeton, Jonathan Marcus, Thomas Conerly, Joshua Baston, and Chris Olah.
\newblock Sparse crosscoders for cross-layer features and model diffing.
\newblock \emph{Transformer Circuits Thread}, 2024.

\bibitem[Marks et~al.(2025)Marks, Rager, Michaud, Belinkov, Bau, and Mueller]{marks2024sparse}
Samuel Marks, Can Rager, Eric~J Michaud, Yonatan Belinkov, David Bau, and Aaron Mueller.
\newblock Sparse feature circuits: Discovering and editing interpretable causal graphs in language models.
\newblock In \emph{The Thirteenth International Conference on Learning Representations}, 2025.
\newblock URL \url{https://openreview.net/forum?id=I4e82CIDxv}.

\bibitem[Mikolov et~al.(2013)Mikolov, Yih, and Zweig]{mikolov-etal-2013-linguistic}
Tomas Mikolov, Wen-tau Yih, and Geoffrey Zweig.
\newblock Linguistic regularities in continuous space word representations.
\newblock In Lucy Vanderwende, Hal Daum{\'e}~III, and Katrin Kirchhoff, editors, \emph{Proceedings of the 2013 Conference of the North {A}merican Chapter of the Association for Computational Linguistics: Human Language Technologies}, pages 746--751, Atlanta, Georgia, June 2013. Association for Computational Linguistics.
\newblock URL \url{https://aclanthology.org/N13-1090/}.

\bibitem[Olah et~al.(2020)Olah, Cammarata, Schubert, Goh, Petrov, and Carter]{olah2020zoom}
Chris Olah, Nick Cammarata, Ludwig Schubert, Gabriel Goh, Michael Petrov, and Shan Carter.
\newblock Zoom in: An introduction to circuits.
\newblock \emph{Distill}, 2020.
\newblock \doi{10.23915/distill.00024.001}.
\newblock https://distill.pub/2020/circuits/zoom-in.

\bibitem[Olshausen and Field(1996)]{olshausen1996emergence}
Bruno~A Olshausen and David~J Field.
\newblock Emergence of simple-cell receptive field properties by learning a sparse code for natural images.
\newblock \emph{Nature}, 381\penalty0 (6583):\penalty0 607--609, 1996.

\bibitem[Oozeer et~al.(2025)Oozeer, Nathawani, Prakash, Lan, Harrasse, and Abdullah]{oozeer2025activation}
Narmeen Oozeer, Dhruv Nathawani, Nirmalendu Prakash, Michael Lan, Abir Harrasse, and Amirali Abdullah.
\newblock Activation space interventions can be transferred between large language models.
\newblock \emph{arXiv preprint arXiv:2503.04429}, 2025.

\bibitem[Panickssery et~al.(2023)Panickssery, Gabrieli, Schulz, Tong, Hubinger, and Turner]{panickssery2023steering}
Nina Panickssery, Nick Gabrieli, Julian Schulz, Meg Tong, Evan Hubinger, and Alexander~Matt Turner.
\newblock Steering llama 2 via contrastive activation addition.
\newblock \emph{arXiv preprint arXiv:2312.06681}, 2023.

\bibitem[Park et~al.(2024)Park, Choe, and Veitch]{park2023linear}
Kiho Park, Yo~Joong Choe, and Victor Veitch.
\newblock The linear representation hypothesis and the geometry of large language models.
\newblock In \emph{Forty-first International Conference on Machine Learning}, 2024.
\newblock URL \url{https://openreview.net/forum?id=UGpGkLzwpP}.

\bibitem[Paulo and Belrose(2025)]{paulo2025sparse}
Gon{\c{c}}alo Paulo and Nora Belrose.
\newblock Sparse autoencoders trained on the same data learn different features.
\newblock \emph{arXiv preprint arXiv:2501.16615}, 2025.

\bibitem[Radford et~al.(2019)Radford, Wu, Child, Luan, Amodei, Sutskever, et~al.]{radford2019language}
Alec Radford, Jeffrey Wu, Rewon Child, David Luan, Dario Amodei, Ilya Sutskever, et~al.
\newblock Language models are unsupervised multitask learners.
\newblock \emph{OpenAI blog}, 1\penalty0 (8):\penalty0 9, 2019.

\bibitem[Raghu et~al.(2017)Raghu, Gilmer, Yosinski, and Sohl-Dickstein]{raghu2017svcca}
Maithra Raghu, Justin Gilmer, Jason Yosinski, and Jascha Sohl-Dickstein.
\newblock Svcca: Singular vector canonical correlation analysis for deep learning dynamics and interpretability.
\newblock \emph{Advances in neural information processing systems}, 30, 2017.

\bibitem[Rajamanoharan et~al.(2024)Rajamanoharan, Lieberum, Sonnerat, Conmy, Varma, Kram{\'a}r, and Nanda]{rajamanoharan2024jumping}
Senthooran Rajamanoharan, Tom Lieberum, Nicolas Sonnerat, Arthur Conmy, Vikrant Varma, J{\'a}nos Kram{\'a}r, and Neel Nanda.
\newblock Jumping ahead: Improving reconstruction fidelity with jumprelu sparse autoencoders.
\newblock \emph{arXiv preprint arXiv:2407.14435}, 2024.

\bibitem[Saphra and Wiegreffe(2024)]{saphra2024mechanistic}
Naomi Saphra and Sarah Wiegreffe.
\newblock Mechanistic?
\newblock \emph{arXiv preprint arXiv:2410.09087}, 2024.

\bibitem[Serre(2006)]{serre2006learning}
Thomas Serre.
\newblock Learning a dictionary of shape-components in visual cortex: comparison with neurons, humans and machines, 2006.

\bibitem[Sharkey et~al.(2025)Sharkey, Chughtai, Batson, Lindsey, Wu, Bushnaq, Goldowsky-Dill, Heimersheim, Ortega, Bloom, et~al.]{sharkey2025open}
Lee Sharkey, Bilal Chughtai, Joshua Batson, Jack Lindsey, Jeff Wu, Lucius Bushnaq, Nicholas Goldowsky-Dill, Stefan Heimersheim, Alejandro Ortega, Joseph Bloom, et~al.
\newblock Open problems in mechanistic interpretability.
\newblock \emph{arXiv preprint arXiv:2501.16496}, 2025.

\bibitem[Stolfo et~al.(2024)Stolfo, Wu, Gurnee, Belinkov, Song, Sachan, and Nanda]{stolfo2024confidence}
Alessandro Stolfo, Ben~Peng Wu, Wes Gurnee, Yonatan Belinkov, Xingyi Song, Mrinmaya Sachan, and Neel Nanda.
\newblock Confidence regulation neurons in language models.
\newblock In \emph{The Thirty-eighth Annual Conference on Neural Information Processing Systems}, 2024.
\newblock URL \url{https://openreview.net/forum?id=0og7nmvDbe}.

\bibitem[Stolfo et~al.(2025)Stolfo, Balachandran, Yousefi, Horvitz, and Nushi]{stolfo2024improving}
Alessandro Stolfo, Vidhisha Balachandran, Safoora Yousefi, Eric Horvitz, and Besmira Nushi.
\newblock Improving instruction-following in language models through activation steering.
\newblock In \emph{The Thirteenth International Conference on Learning Representations}, 2025.
\newblock URL \url{https://openreview.net/forum?id=wozhdnRCtw}.

\bibitem[Team et~al.(2024)Team, Riviere, Pathak, Sessa, Hardin, Bhupatiraju, Hussenot, Mesnard, Shahriari, Ram{\'e}, et~al.]{team2024gemma}
Gemma Team, Morgane Riviere, Shreya Pathak, Pier~Giuseppe Sessa, Cassidy Hardin, Surya Bhupatiraju, L{\'e}onard Hussenot, Thomas Mesnard, Bobak Shahriari, Alexandre Ram{\'e}, et~al.
\newblock Gemma 2: Improving open language models at a practical size.
\newblock \emph{arXiv preprint arXiv:2408.00118}, 2024.

\bibitem[Todd et~al.(2024)Todd, Li, Sharma, Mueller, Wallace, and Bau]{todd2023function}
Eric Todd, Millicent Li, Arnab~Sen Sharma, Aaron Mueller, Byron~C Wallace, and David Bau.
\newblock Function vectors in large language models.
\newblock In \emph{The Twelfth International Conference on Learning Representations}, 2024.
\newblock URL \url{https://openreview.net/forum?id=AwyxtyMwaG}.

\bibitem[Wei et~al.(2022)Wei, Tay, Bommasani, Raffel, Zoph, Borgeaud, Yogatama, Bosma, Zhou, Metzler, Chi, Hashimoto, Vinyals, Liang, Dean, and Fedus]{wei2022emergent}
Jason Wei, Yi~Tay, Rishi Bommasani, Colin Raffel, Barret Zoph, Sebastian Borgeaud, Dani Yogatama, Maarten Bosma, Denny Zhou, Donald Metzler, Ed~H. Chi, Tatsunori Hashimoto, Oriol Vinyals, Percy Liang, Jeff Dean, and William Fedus.
\newblock Emergent abilities of large language models.
\newblock \emph{Transactions on Machine Learning Research}, 2022.
\newblock ISSN 2835-8856.
\newblock URL \url{https://openreview.net/forum?id=yzkSU5zdwD}.
\newblock Survey Certification.

\bibitem[Zhou et~al.(2023)Zhou, Lu, Mishra, Brahma, Basu, Luan, Zhou, and Hou]{zhou2023ifeval}
Jeffrey Zhou, Tianjian Lu, Swaroop Mishra, Siddhartha Brahma, Sujoy Basu, Yi~Luan, Denny Zhou, and Le~Hou.
\newblock Instruction-following evaluation for large language models, 2023.
\newblock URL \url{https://arxiv.org/abs/2311.07911}.

\bibitem[Zou et~al.(2023)Zou, Wang, Carlini, Nasr, Kolter, and Fredrikson]{zou2023universal}
Andy Zou, Zifan Wang, Nicholas Carlini, Milad Nasr, J~Zico Kolter, and Matt Fredrikson.
\newblock Universal and transferable adversarial attacks on aligned language models.
\newblock \emph{arXiv preprint arXiv:2307.15043}, 2023.

\end{thebibliography}

\newpage
\newpage

\appendix
\section{Stitch Training Details}
The stitch training methodology is quite minimal. For the training dataset, we collect activations at the desired layers in both models over the first $180$k samples of OpenWebText with a context size of $512$ ($128$ for Gemma due to compute constraints) and evaluate over the next $1$k samples with the same context size. We mask out all special tokens. The stitches themselves are two separate Linear layers (with biases initialized to $0$) that map between the dimensions of $A$ and $B$. We use the Adam optimizer with a learning rate of \texttt{1e-4} and clip gradient norms to $1.0$. We found minimal sensitivity to learning rate schedule, so we just use a cosine annealing decay, and found that $2$ epochs is sufficient for convergence (though even $1$ is probably enough).

\subsection{Layer Selection}\label{sec:SVCCA}
\begin{figure}[ht]
    \centering
    \begin{subfigure}[b]{0.45\linewidth}
        \includegraphics[width=\linewidth]{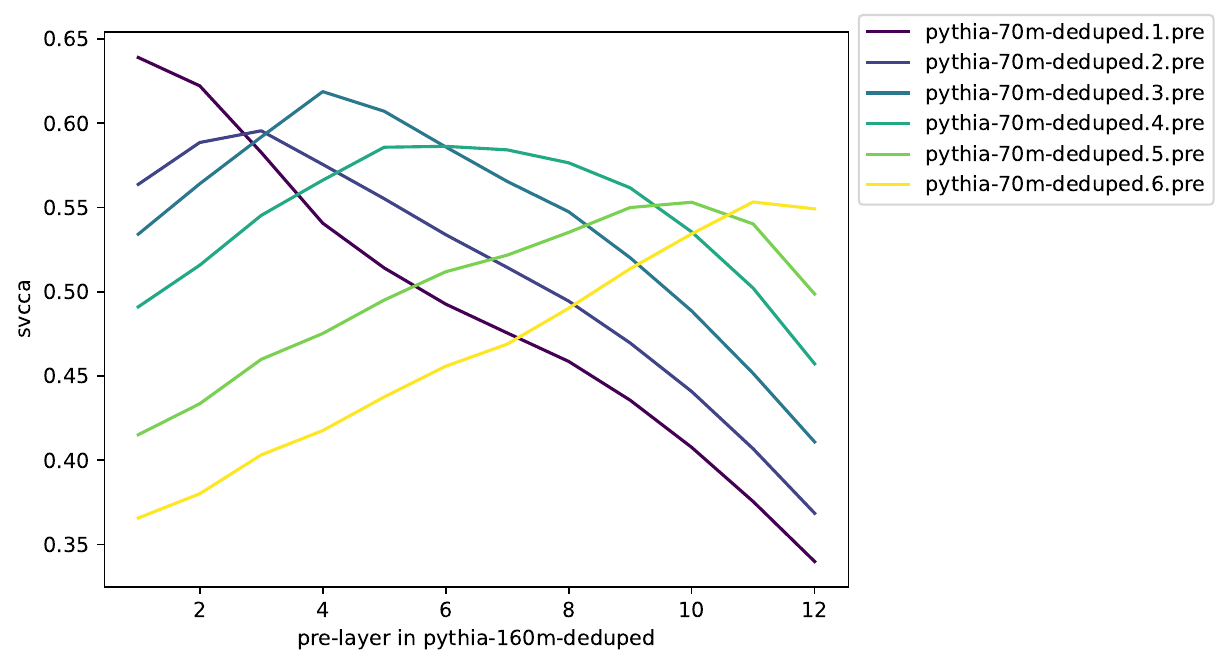}
        \caption{pythia-70m to pythia-160m}
    \end{subfigure}
    \hfill
    \begin{subfigure}[b]{0.45\linewidth}
        \includegraphics[width=\linewidth]{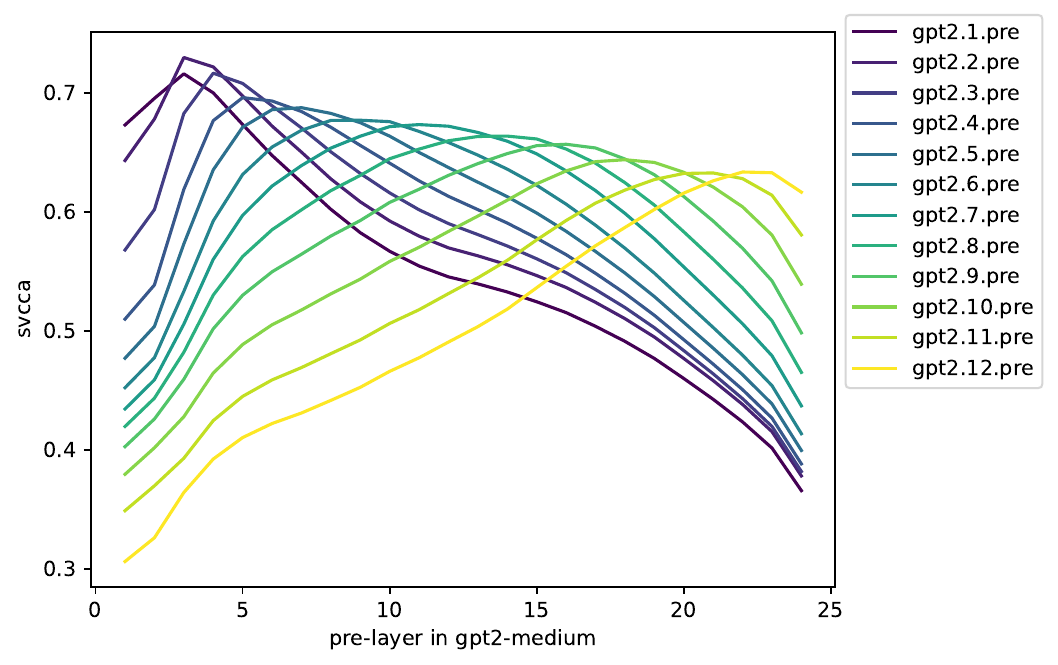}
        \caption{gpt2-small to gpt2-medium}
    \end{subfigure}
    \hfill
    \begin{subfigure}[b]{0.45\linewidth}
        \includegraphics[width=\linewidth]{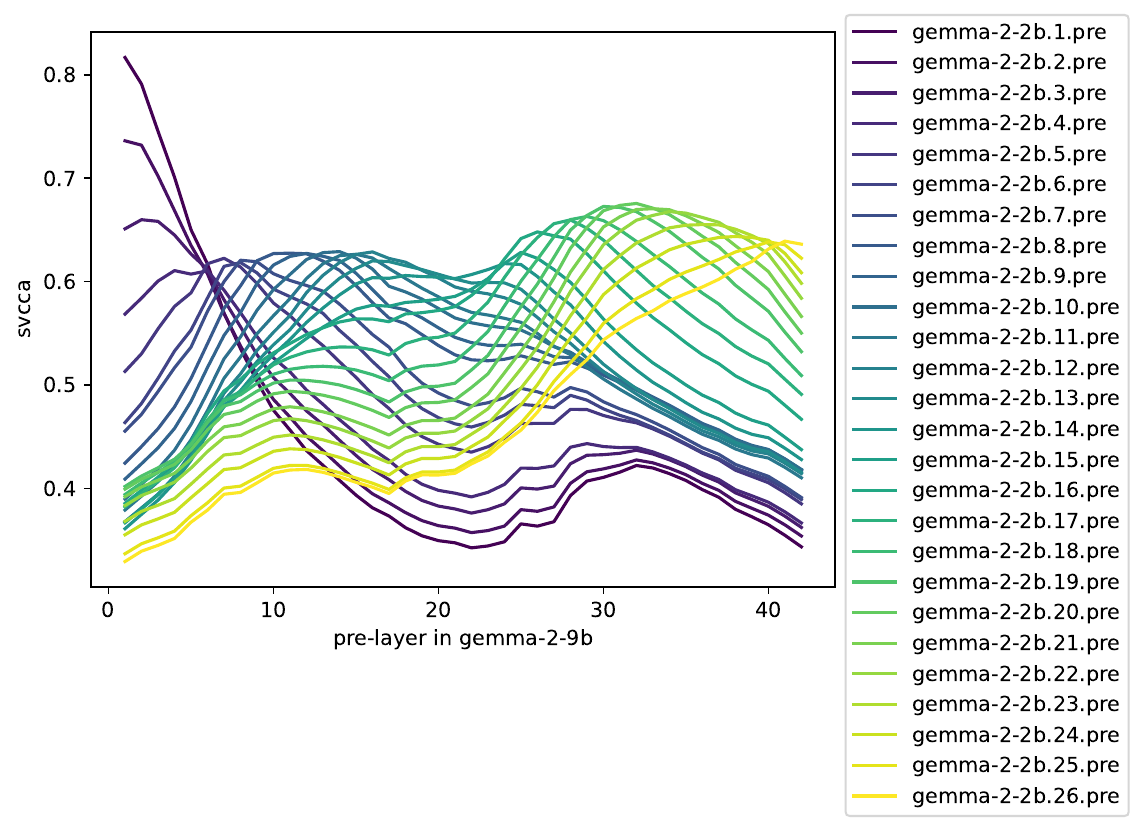}
        \caption{gemma-2-2b to gemma-2-9b}
    \end{subfigure}
    \caption{Computed SVCCA scores over all pairwise layers in the model pairs we stitch between. We end arbitrarily selecting a layer in the smaller model and choosing the layer in $B$ with the highest SVCCA coefficient.}
    \label{fig:svccas}
\end{figure}

Suppose we have fixed some layer $\ell_A$ in model $A$ that we would like to stitch from. We determine the layer $\ell_B$ by computing a Singular Vector Canonical Correlation Analysis (SVCCA) over a small sample of tokens between the residual stream activations at $\ell_A$ and the candidate layer $\ell_B$ and taking the argmax. We use SVCCA because it is cheap to over a small set of activations while being directly related to how linearly related the two sets of activations accounting for noise in lower variance directions. We visualize all pairwise SVCCA values computed in Figure \ref{fig:svccas}.
\subsection{Inversion Ablation}\label{sec:inversion_ablation}
\begin{figure}[ht]
    \centering
    \includegraphics[width=0.75\linewidth]{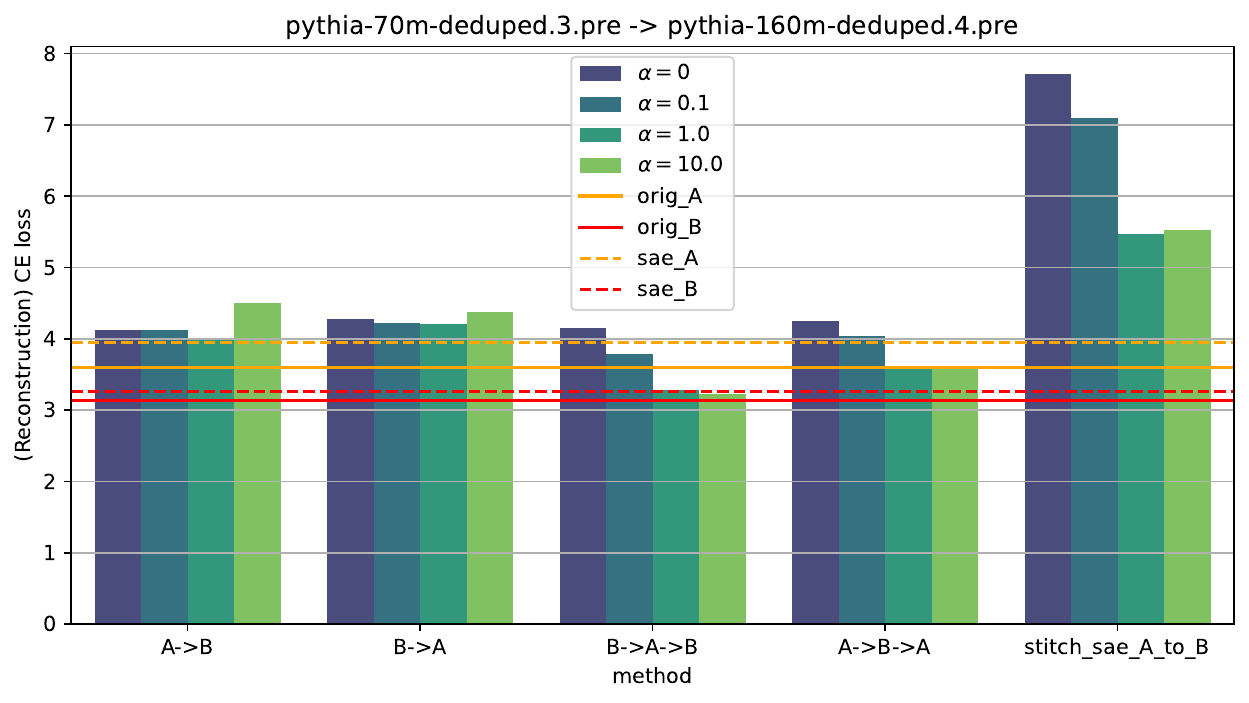}
    \caption{Strength of inversion ablation experiment. For various values of $\alpha$ we plot downstream next token CE loss for different transformations that should all be low - stitching from $A$ to $B$ and from $B$ to $A$, inversions, and the delta loss of an SAE transferred up from $A$ to $B$. We can see that $\alpha = 1.0$ strikes a good balance so we use it for all experiments.}
    \label{fig:inversion_ablation}
\end{figure}

See Figure \ref{fig:inversion_ablation}.

\section{Transferring Details}
Assume, we have two autoregressive language models $A$ and $B$ with latent embedding dimensions $d_A$ and $d_B$, respectively. Furthermore, assume there exist affine transformations $\mathcal{T}_\uparrow: (P_\uparrow, b_\uparrow): \mathbb{R}^{d_A \times d_B} \times \mathbb{R}^{d_B}$ and $\mathcal{T}_\downarrow = (P_\downarrow, b_\downarrow) : \mathbb{R}^{d_B \times d_A} \times \mathbb{R}^{d_A}$ as in Equation \ref{eq:affine_Ts} that relate the residual streams of $A$ and $B$ at fixed layers in both models.

\subsection{SAE Methodology}\label{sec:transferring_sae_appendix}
Recall that a $\text{SAE}: \mathbb{R}^{d_A} \to \mathbb{R}^{d_A}$ defined on model $A$ can be written using the following computation:
\begin{align}
    \text{SAE}(x; \theta) = \sigma(xW_e + b_e)W_d + b_d,
\end{align}
parameterized by the $4$-tuple $\theta = (W_e, b_e, W_d, b_d)$ and an activation function $\sigma$.

The insight is that we can think of transferring a sparse autoencoder from model $A$ to model $B$ as capturing the following computation. For every latent in model $B$ $h_B \in \mathbb{R}^{d_B}$: 
\begin{enumerate}
    \item Stitch $h_B$ to $A$ using $\mathcal{T}_\downarrow$.
    \item Apply the SAE on the stitched latent.
    \item Stitch the reconstruction back to $B$ using $\mathcal{T}_\uparrow$.
\end{enumerate}
It turns out this computation can be collapsed into a reparameterized SAE on model $B$ because all of the transformations are affine.
\begin{align*}
    (\mathcal{T}_\uparrow \circ \textbf{SAE} \circ \mathcal{T}_\downarrow) (h_B) &= \textbf{SAE}(h_B P_\downarrow + b_\downarrow; \theta) P_\uparrow + b_\uparrow\\
    &= \Big(\sigma([h_B P_\downarrow + b_\downarrow] W_e + b_e) W_d + b_d\Big) P_\uparrow + b_\uparrow \\
    &= \sigma(h_B [P_\downarrow W_e] + [b_\downarrow W_e + b_e])[W_dP_\uparrow] + [b_dP_\uparrow + b_\uparrow]\\
    &= \textbf{SAE}(h_B; \theta'),
\end{align*}
where \begin{equation}
    \theta' = (P_\downarrow W_e, b_\downarrow W_e + b_e, W_dP_\uparrow, b_dP_\uparrow + b_\uparrow)
\end{equation}
is another sparse autoencoder, but now mapping from $\mathbb{R}^{d_B} \to \mathbb{R}^{d_B}$. Remark: one slight caveat is that $\text{SAE}'$'s feature matrices are still only rank $d_A$.

Observe that the matrices $P_\uparrow$ and $P_\downarrow$ do the grunt work in manipulating the actual feature spaces, whereas the biases $b_\uparrow$ and $b_\downarrow$ are just relocating the ``position'' of the residual stream in space and do not adjust the feature spaces themselves.

We also notice that assuming $\textbf{SAE}(x; \theta) \approx x$, the reconstruction loss \begin{equation}
    \|\textbf{SAE}(h_B; \theta') - h_B\|_2
\end{equation} is low when $\mathcal{T}_\uparrow$ and $\mathcal{T}_\downarrow$ invert each other (as expected). This inspires the inclusion of the additional inversion penalty in training the transformations $\mathcal{T}_\uparrow$ and $\mathcal{T}_\downarrow$ (Equation \ref{eq:stitching_loss_objective}).
\subsection{Probes and Steering Vectors}
To transfer a linear probe from $A$ to $B$, we simply apply the probe on the down-stitched residual stream $\mathcal{T}_\downarrow(B^{\ell_B}(t))$. If the probe is linear and has normal direction $w$, it is equivalent to using the vector $w P_\downarrow^T$ in $B$. Steering vectors are just transferred directly using $P_\uparrow$ without the bias.
\section{Zero-Shot SAE Evaluations}\label{sec:zero_shot_evaluations}
As basic evaluations of how well the feature spaces match, we can compute some core metrics of our zero shot transferred SAEs and compare them against the original SAEs. We report $L_0$, Fraction of Unexplained Variance (FUV), Delta loss, and dead features $\%$ in Table \ref{tab:zero_shot_sae_metrics}. Importantly, we note that the transferred SAEs are not perfect - one explanation of why is that the weight matrices of the transferred SAE (Equation \ref{eq:transferred_sae}) are still rank $\leq d_A$.
\begin{table}[]
    \centering
    \caption{Summary metrics of transferred SAEs (no training). Results are displayed as original (A) / transfer (B). $L_0$ evaluations are dependent on architecture - the Pythia and GPT2 SAEs are Top-K and Gemma are JumpReLU.}
    \resizebox{\textwidth}{!}{\begin{tabular}{lccc}
    \toprule
         Metric & pythia-70m.3 / pythia-160m.4 & gpt2-small.6 / gpt2-medium.10 & gemma-2-2b.20 / gemma-2-9b.33 \\
         \midrule
         $L_0$ & 16.0 / 16.0  & 32.0 / 32.0 & 77.2 / 74.7 \\
         FUV & 0.17 / 0.52 & 0.11 / 0.39 & 0.21 / 0.42 \\
         Delta Loss  & 0.36 / 2.16 & 0.08 / 0.96 &  0.50 / 0.85 \\
         Dead Features \%  & 5.2\% / 2.9\% & <1\% / < 1\% &  1.5\% / 3.5\% \\
         \bottomrule
    \end{tabular}}
    \label{tab:zero_shot_sae_metrics}
\end{table}
\section{SAE Training and FLOP Estimation Details}\label{sec:sae_training_and_flop_estimation}
All SAEs we train are TopK SAEs trained using SAELens on unnormalized residual stream activations. We train SAEs with latent sizes $4096$, $8192$, $16384$, $32768$, and $65536$. We abide by the following practices:
\begin{enumerate}
    \item We normalize the decoder vectors to unit norm each iteration.
    \item When randomly initializing, we initialize the decoder and encoder as transposes of each other.
    \item We use a constant learning rate schedule and just use $0.0001$ as the learning rate.
    \item We do not use an auxiliary loss for ease of FLOPs estimation (discussed below).
\end{enumerate}

SAELens includes the auxiliary loss for TopK training by default. In order to work around disabling it, we set the \texttt{dead\_feature\_window} parameter to an arbitrarily large number so that dead feature computations are never run during training.

We estimate FLOP counts of training the stitches and SAEs with cached activations by simply computing the FLOPs of one forward and backward pass for the desired module on a batch of dummy data inside of the context manager \texttt{torch.utils.flop\_counter.FlopCounterMode}. We then scale by the number of training iterations. Example FLOP estimates for a set of SAEs is shown in \ref{tab:efficient_sae_results_flops}. 
\begin{table}[h]
    \centering
    \caption{FLOPs estimates for various important procedures with caching activations. All SAEs are trained with sparsity $k=64$ and width $32$k. A full run is $4$B tokens ($120$k iterations) for the SAEs and $200$M ($36$k iterations) tokens for the stitch. The Pythia-70m SAE is trained post layer $2$ and the Pythia-160m SAEs are trained post layer $3$.}
    \begin{tabular}{ll}
         Procedure & FLOPs (w/ caching)\\
         \midrule
         Pythia-70m SAE (Scratch) & $8.2 \times 10^{16}$ \\
         Pythia-160m SAE (Scratch \& Transfer) & $1.2 \times 10^{17}$ \\
         Stitching Layer & $1.4 \times 10^{15}$ \\
    \end{tabular}
    \label{tab:efficient_sae_results_flops}
\end{table}
As a minor fact, we do find that with larger latent sizes, the number of dead features in the \textit{original} SAE (on $A$) increases since we are not using the auxiliary loss. Since dead features tend to be inherited in the stitched SAE initialization, we end up with more dead features when training from the stitched SAE initialization. This could be remediated by reincluding the auxiliary loss (which would require careful FLOPs estimation), but we did not test this claim. 
\subsection{GPT2 SAE Transfer}\label{sec:gpt2_sae_transfer}
\begin{figure}
    \centering
    \begin{subfigure}[b]{0.45\linewidth}
    \includegraphics[width=\linewidth]{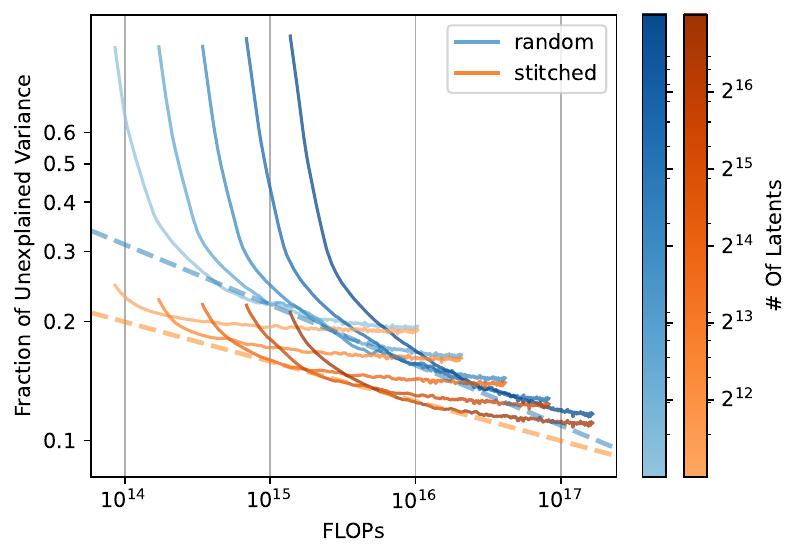}
    \caption{}
    \label{fig:gpt2_sae_scaling_law}
    \end{subfigure}
    \hfill
    \begin{subfigure}[b]{0.45\linewidth}
    \includegraphics[width=\linewidth]{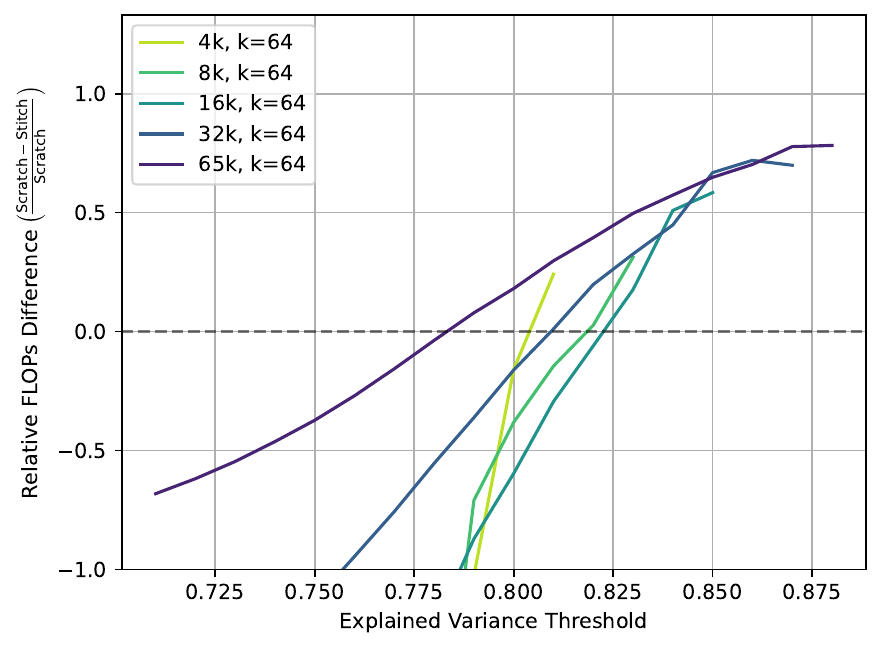}
    \caption{}
    \label{fig:gpt2_relative_flops_diff}
    \end{subfigure}
    \caption{(a) SAE ``scaling law'' for transfer results on GPT2 models with $k = 32$. (b) Relative FLOPs Difference between the initializations.}
    \label{fig:gpt2_sae_transfer}
\end{figure}
In Figure \ref{fig:gpt2_sae_transfer}, we replicate the same plots as Figure \ref{fig:fig1}c but for the GPT2 series models, with SAEs trained on 2B tokens and $k = 32$. 

\subsection{Scaling Law Fit}\label{sec:scaling_law_fitting}
We fit the scaling law by sampling the frontier MSE achieved at various FLOPs thresholds and fitting a linear model on the log-log scale (as opposed to fitting an irreducible loss term - unfortunately we do not have enough high-FLOPs points due to compute constraints to fit a complete scaling law stably). However, for our purposes, since theoretically the irreducible loss is the same for all initializations, so purely for relative comparison between two scaling laws it is not crucial. This results in a law of the form 
\begin{equation}
    L(C) \approx AC^{-\beta}
\end{equation}
instead of \begin{equation}
    L(C) \approx L_\infty + AC^{-\beta}
\end{equation}
where $L_\infty$ is the irreducible loss. We report the fitted coefficients for the approximate laws in Figure \ref{fig:fig1}c (Pythia) and Figure \ref{fig:gpt2_sae_scaling_law} (GPT2) in Table \ref{tab:fitted_law_coefficients}.

\begin{table}[h]
\centering
\caption{Approximate scaling law fitted coefficients for all laws we fit.}
\begin{tabular}{lcc}
\toprule
Model/[Scratch, Stitch]     & $A$  & $\beta$ \\\midrule
pythia-160m-deduped/scratch & 41.2 & 0.16    \\
pythia-160m-deduped/stitch  & 18.5 & 0.14    \\
gpt2-medium/scratch            & 42.3 & 0.15    \\
gpt2-medium/stitch             & 5.0  & 0.10   \\
\bottomrule
\end{tabular}
\label{tab:fitted_law_coefficients}
\end{table}

\section{Full Probing Results}\label{sec:full_probing_results}
\begin{figure}
    \centering
    \includegraphics[width=0.9\linewidth]{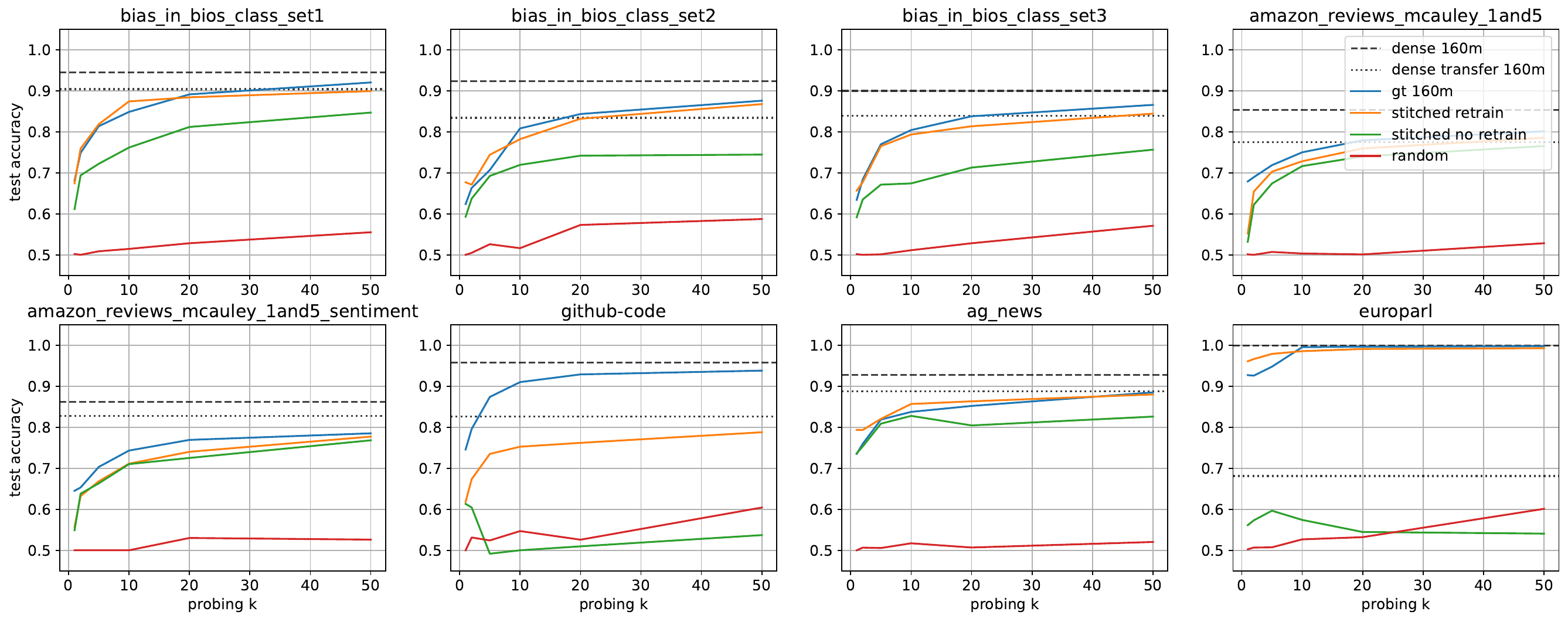}
    \caption{Transferring sparse probes from pythia-70m-deduped to pythia-160m-deduped works reasonably well in most datasets except code and language without retraining the probe on pythia-160m activations. Retraining the probe on the same features recovers ground truth performance.}
    \label{fig:retrain_probing_results}
\end{figure}

\begin{figure}
    \centering
    \includegraphics[width=0.9\linewidth]{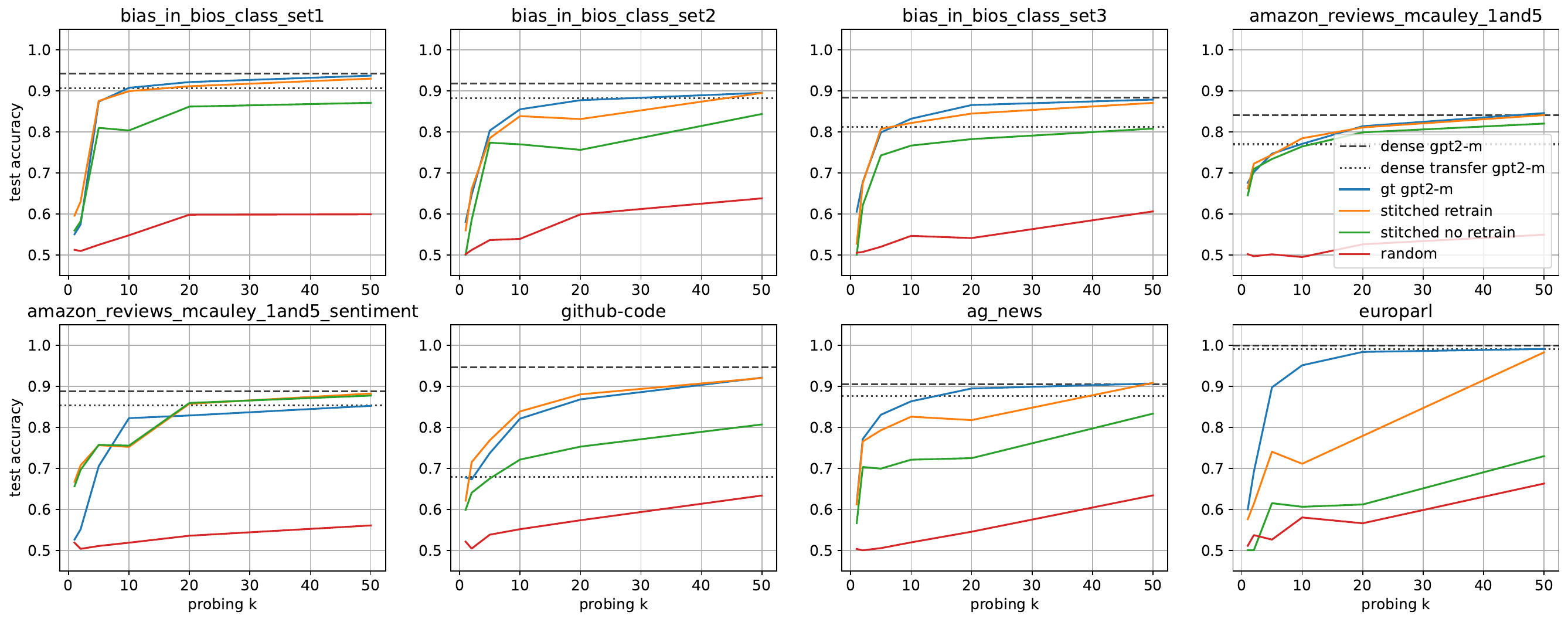}
    \caption{Sparse probing results on GPT2 SAEs trained from scratch. The conclusions are similar to the Pythia results - we can probe better than random using transferred probes. Code and language domains again seem hard to directly transfer but are fixed with retraining the probes.}
    \label{fig:sparse_probing_gpt2}
\end{figure}

See Figure \ref{fig:retrain_probing_results} and Figure \ref{fig:sparse_probing_gpt2} for the fully decomposed probing results on each of the $8$ SAEBench datasets.
\section{Steering Experiment Details}\label{sec:steering_experiment_details}
We do feature selection and coefficient fitting over the first $250$ samples from Europarl for the dataset. To evaluate, we use \texttt{langdetect} to detect the most likely language of the response to the input prompt. We compute confidence intervals using Clopper-Pearson intervals at the $\alpha = 0.05$ level.
\begin{table}[]
\centering
\caption{Full language steering results for gemma-2-2b. Format is [accuracy, (confidence interval low, confidence interval high)].}
\label{tab:2b_full_results}
\begin{tabular}{@{}cccc@{}}
\toprule
\textbf{}      & \textbf{No Steering (2b)} & \textbf{Steering (2b)} & \textbf{Transfer Steering (9b$\to$2b)} \\ \midrule
\textbf{bg-en} & 0.00 (0.00, 0.02)    & 0.31 (0.24, 0.38)      & 0.01 (0.00, 0.03)                      \\
\textbf{cs-en} & 0.00 (0.00, 0.02)    & 0.42 (0.34, 0.50)      & 0.02 (0.00, 0.05)                      \\
\textbf{da-en} & 0.01 (0.00, 0.04)    & 0.42 (0.34, 0.50)      & 0.25 (0.18, 0.32)                      \\
\textbf{de-en} & 0.00 (0.00, 0.02)    & 0.81 (0.74, 0.87)      & 0.79 (0.71, 0.85)                      \\
\textbf{el-en} & 0.00 (0.00, 0.02)    & 0.85 (0.79, 0.90)      & 0.48 (0.41, 0.56)                      \\
\textbf{en-es} & 0.01 (0.00, 0.03)    & 0.74 (0.66, 0.80)      & 0.69 (0.61, 0.76)                      \\
\textbf{en-et} & 0.00 (0.00, 0.02)    & 0.28 (0.21, 0.35)      & 0.01 (0.00, 0.04)                      \\
\textbf{en-fi} & 0.00 (0.00, 0.02)    & 0.40 (0.32, 0.48)      & 0.06 (0.03, 0.10)                      \\
\textbf{en-fr} & 0.01 (0.00, 0.03)    & 0.83 (0.76, 0.88)      & 0.77 (0.69, 0.83)                      \\
\textbf{en-hu} & 0.01 (0.00, 0.03)    & 0.72 (0.65, 0.79)      & 0.06 (0.03, 0.10)                      \\
\textbf{en-it} & 0.00 (0.00, 0.02)    & 0.69 (0.62, 0.76)      & 0.60 (0.52, 0.67)                      \\
\textbf{en-lt} & 0.00 (0.00, 0.02)    & 0.23 (0.17, 0.31)      & 0.00 (0.00, 0.02)                      \\
\textbf{en-lv} & 0.00 (0.00, 0.02)    & 0.50 (0.42, 0.58)      & 0.00 (0.00, 0.02)                      \\
\textbf{en-nl} & 0.00 (0.00, 0.02)    & 0.67 (0.59, 0.74)      & 0.64 (0.57, 0.72)                      \\
\textbf{en-pl} & 0.01 (0.00, 0.03)    & 0.62 (0.54, 0.69)      & 0.18 (0.12, 0.25)                      \\
\textbf{en-pt} & 0.01 (0.00, 0.03)    & 0.76 (0.69, 0.82)      & 0.63 (0.55, 0.70)                      \\
\textbf{en-ro} & 0.00 (0.00, 0.02)    & 0.40 (0.32, 0.48)      & 0.09 (0.05, 0.15)                      \\
\textbf{en-sk} & 0.00 (0.00, 0.02)    & 0.19 (0.13, 0.26)      & 0.01 (0.00, 0.04)                      \\
\textbf{en-sl} & 0.00 (0.00, 0.02)    & 0.28 (0.21, 0.35)      & 0.02 (0.00, 0.05)                      \\
\textbf{en-sv} & 0.00 (0.00, 0.02)    & 0.69 (0.62, 0.76)      & 0.48 (0.41, 0.56)                      \\ \bottomrule
\end{tabular}
\end{table}
\begin{table}[]
\centering
\caption{Full language steering results for gemma-2-9b. Format is [accuracy, (confidence interval low, confidence interval high)].}
\label{tab:9b_full_results}
\begin{tabular}{@{}cccc@{}}
\toprule
               & \textbf{No Steering (9b)} & \textbf{Steering (9b)} & \textbf{Transfer Steering (2b$\to$9b)} \\
               \midrule
\textbf{bg-en} & 0.00 (0.00, 0.02)         & 0.24 (0.18, 0.31)      & 0.00 (0.00, 0.02)                      \\
\textbf{cs-en} & 0.00 (0.00, 0.02)         & 0.55 (0.47, 0.63)      & 0.00 (0.00, 0.02)                      \\
\textbf{da-en} & 0.00 (0.00, 0.02)         & 0.65 (0.57, 0.72)      & 0.14 (0.09, 0.20)                      \\
\textbf{de-en} & 0.00 (0.00, 0.02)         & 0.85 (0.78, 0.90)      & 0.83 (0.77, 0.89)                      \\
\textbf{el-en} & 0.00 (0.00, 0.02)         & 0.87 (0.81, 0.92)      & 0.04 (0.02, 0.09)                      \\
\textbf{en-es} & 0.01 (0.00, 0.03)         & 0.80 (0.73, 0.86)      & 0.69 (0.61, 0.76)                      \\
\textbf{en-et} & 0.00 (0.00, 0.02)         & 0.36 (0.29, 0.44)      & 0.00 (0.00, 0.02)                      \\
\textbf{en-fi} & 0.00 (0.00, 0.02)         & 0.17 (0.12, 0.24)      & 0.01 (0.00, 0.04)                      \\
\textbf{en-fr} & 0.01 (0.00, 0.03)         & 0.09 (0.05, 0.15)      & 0.69 (0.61, 0.76)                      \\
\textbf{en-hu} & 0.01 (0.00, 0.03)         & 0.02 (0.00, 0.05)      & 0.01 (0.00, 0.03)                      \\
\textbf{en-it} & 0.00 (0.00, 0.02)         & 0.07 (0.04, 0.13)      & 0.56 (0.48, 0.64)                      \\
\textbf{en-lt} & 0.00 (0.00, 0.02)         & 0.00 (0.00, 0.02)      & 0.00 (0.00, 0.02)                      \\
\textbf{en-lv} & 0.00 (0.00, 0.02)         & 0.00 (0.00, 0.02)      & 0.00 (0.00, 0.02)                      \\
\textbf{en-nl} & 0.00 (0.00, 0.02)         & 0.09 (0.05, 0.14)      & 0.53 (0.45, 0.61)                      \\
\textbf{en-pl} & 0.00 (0.00, 0.02)         & 0.01 (0.00, 0.03)      & 0.01 (0.00, 0.03)                      \\
\textbf{en-pt} & 0.01 (0.00, 0.03)         & 0.09 (0.05, 0.14)      & 0.62 (0.54, 0.69)                      \\
\textbf{en-ro} & 0.00 (0.00, 0.02)         & 0.00 (0.00, 0.02)      & 0.01 (0.00, 0.03)                      \\
\textbf{en-sk} & 0.00 (0.00, 0.02)         & 0.40 (0.33, 0.48)      & 0.01 (0.00, 0.03)                      \\
\textbf{en-sl} & 0.00 (0.00, 0.02)         & 0.27 (0.20, 0.34)      & 0.00 (0.00, 0.02)                      \\
\textbf{en-sv} & 0.00 (0.00, 0.02)         & 0.79 (0.71, 0.85)      & 0.31 (0.24, 0.39)                    \\ 
\bottomrule
\end{tabular}
\end{table}
\subsection{Specific Language Results}\label{sec:specific_language_results}
We report the full accuracies for all languages in Tables \ref{tab:2b_full_results} and \ref{tab:9b_full_results}.

The clipped relative transfer gap is defined as 
\begin{equation}
    \text{Clip}\left(\frac{\text{Tranfer Perf}}{\text{Ground Truth Perf}}, 0, 1\right),
\end{equation}
and we define $0/0$ as $0$ and $\infty$ or $-\infty$ are clipped to $1$ and $0$ respectively.

\begin{figure}
    \centering
    \includegraphics[width=0.45\linewidth]{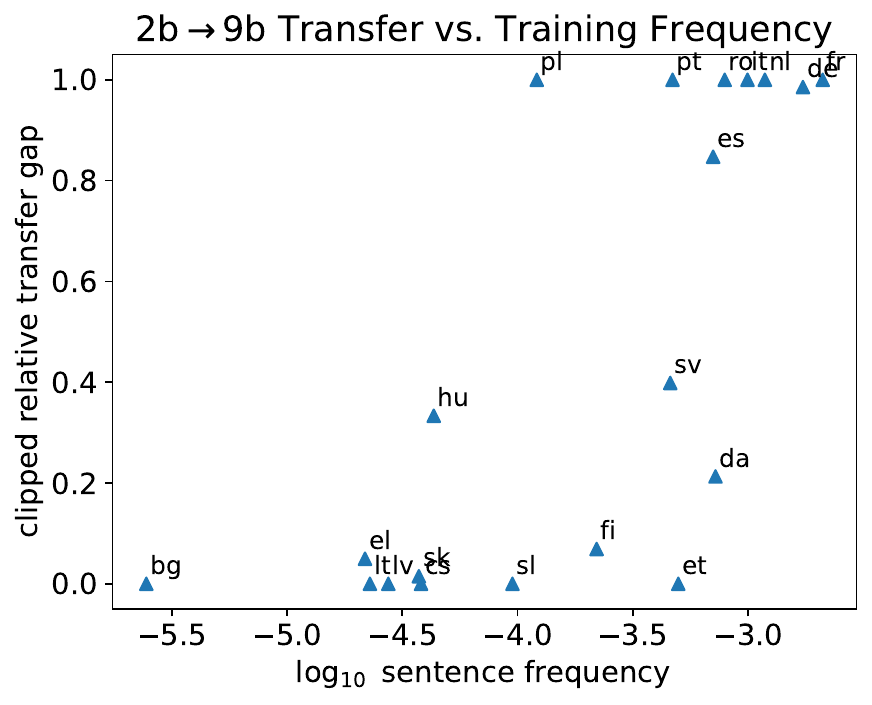}
    \includegraphics[width=0.45\linewidth]{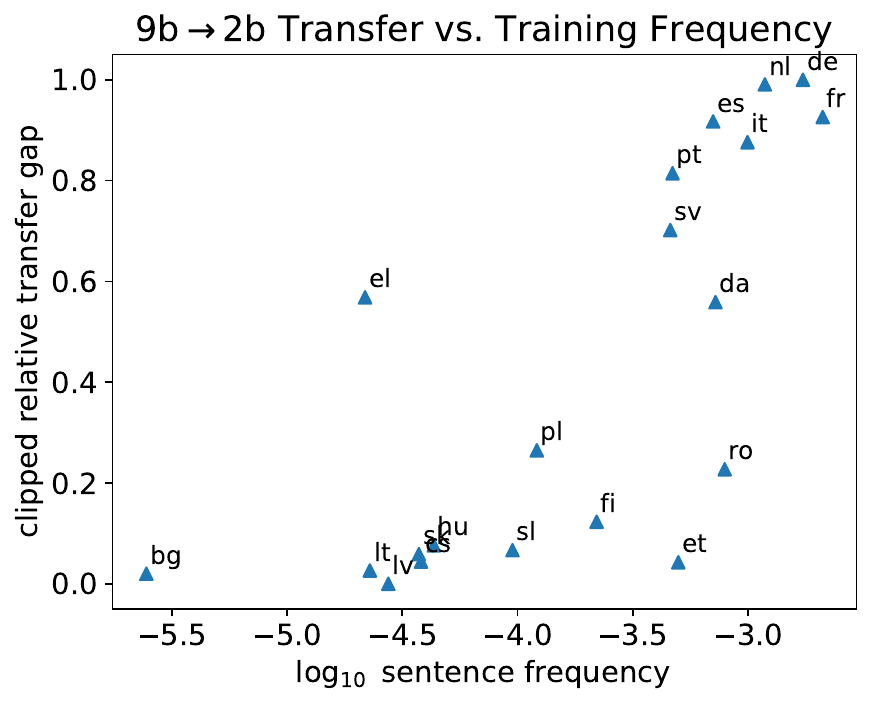}
    \caption{A positive correlation exists between being more frequent in natural language training data and having stronger transfer steering performance (relative to ground truth performance).}
    \label{fig:relative_transfer_gap_vs_frequency}
\end{figure}

We can compute a language frequency approximation by sentence-tokenizing a corpus of text and using \texttt{langdetect} to classify the language of each sentence, counting the frequencies. In Figure \ref{fig:relative_transfer_gap_vs_frequency}, we plot the clipped relative transfer gap against the language log frequency (computed over $100$k samples of OpenWebText) and identify a positive correlation with frequent languages having better transfer performance.
\subsection{Other Steering Experiments}\label{sec:general_steering}
We also trained a stitch between gemma-2-2b-it.20 and gemma-2-9b-it.33 and computed steering vectors for instruction following tasks directly following \citet{stolfo2024improving}. The task setup involves responding to prompts without explicit instructions formatted using chat templating. We then apply a DiffMean steering vector at each token and verify whether the instruction is followed by the response. We found weaker steering results - in no case did transfer steering improve upon ground truth steering and small to big results are weak in general. In some cases, we find that ground truth steering performance is matched and generally stitched steering can still provide improvements over no steering whatsoever in some instructions.
\begin{figure}
    \centering
    \includegraphics[width=0.74\linewidth]{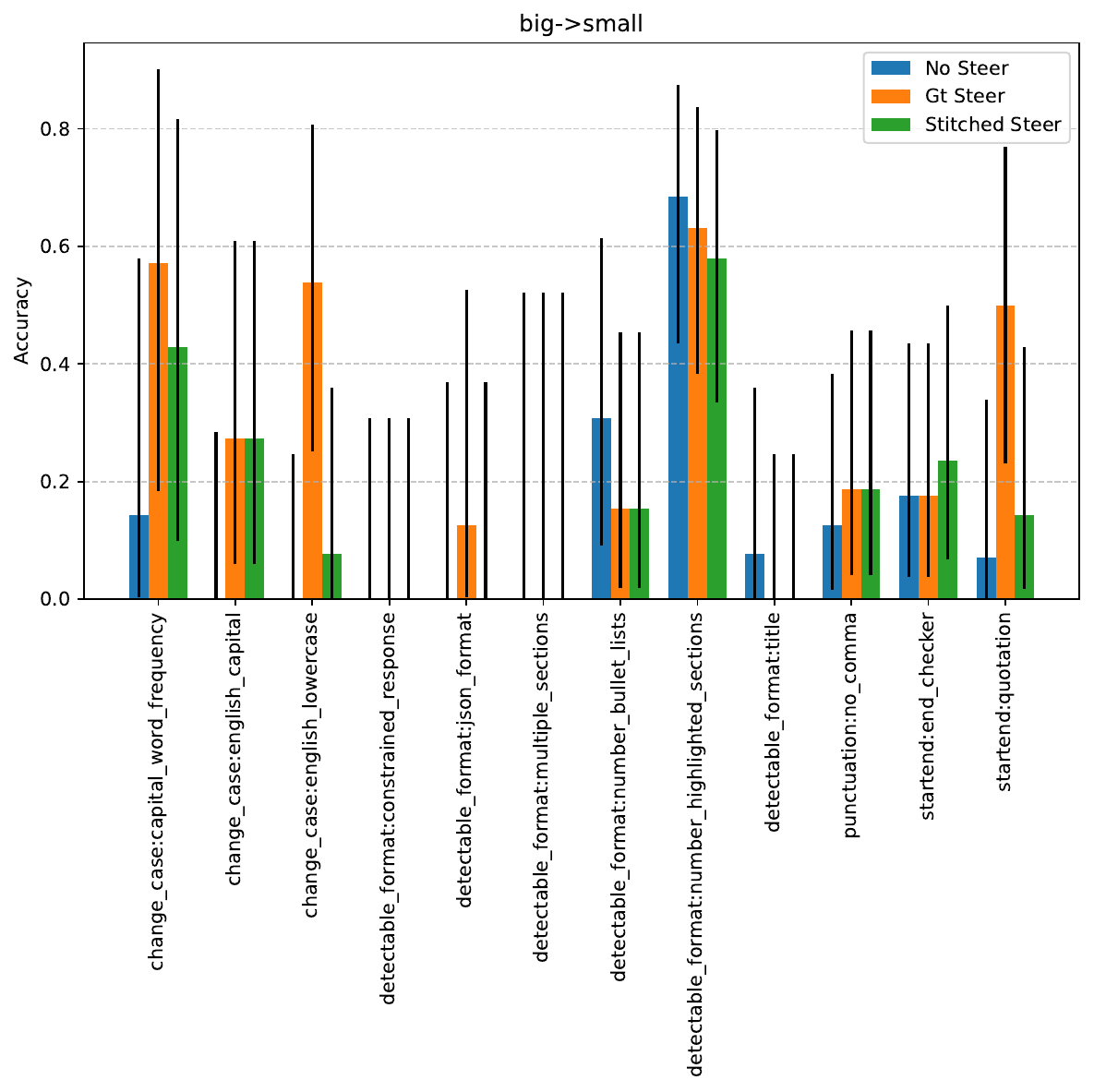}\\
    \includegraphics[width=0.5\linewidth]{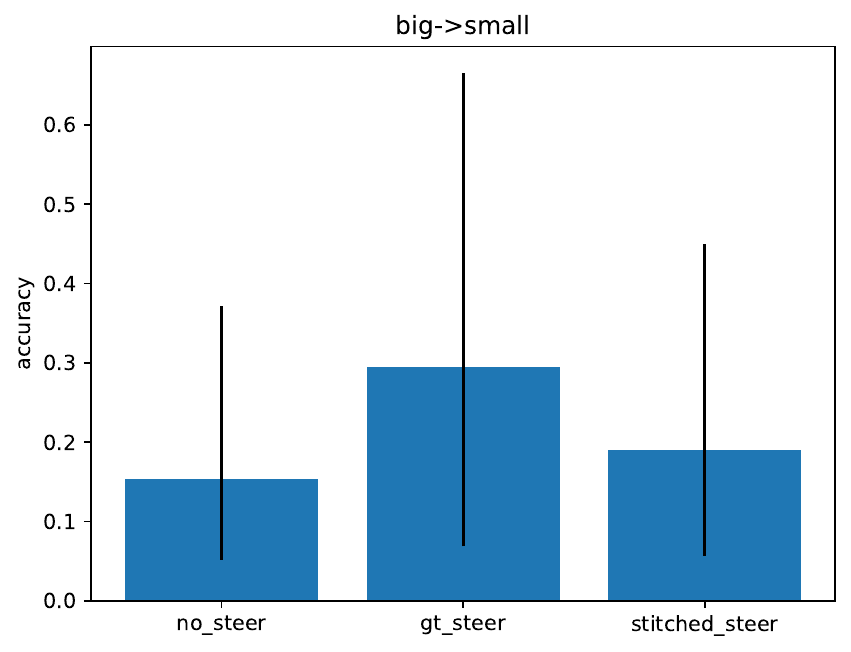}
    \caption{Big to small IFEval steering. We find that casing instructions are generally easiest to transfer.}
    \label{fig:ifeval_big_small}
\end{figure}
\begin{figure}
    \centering
    \includegraphics[width=0.74\linewidth]{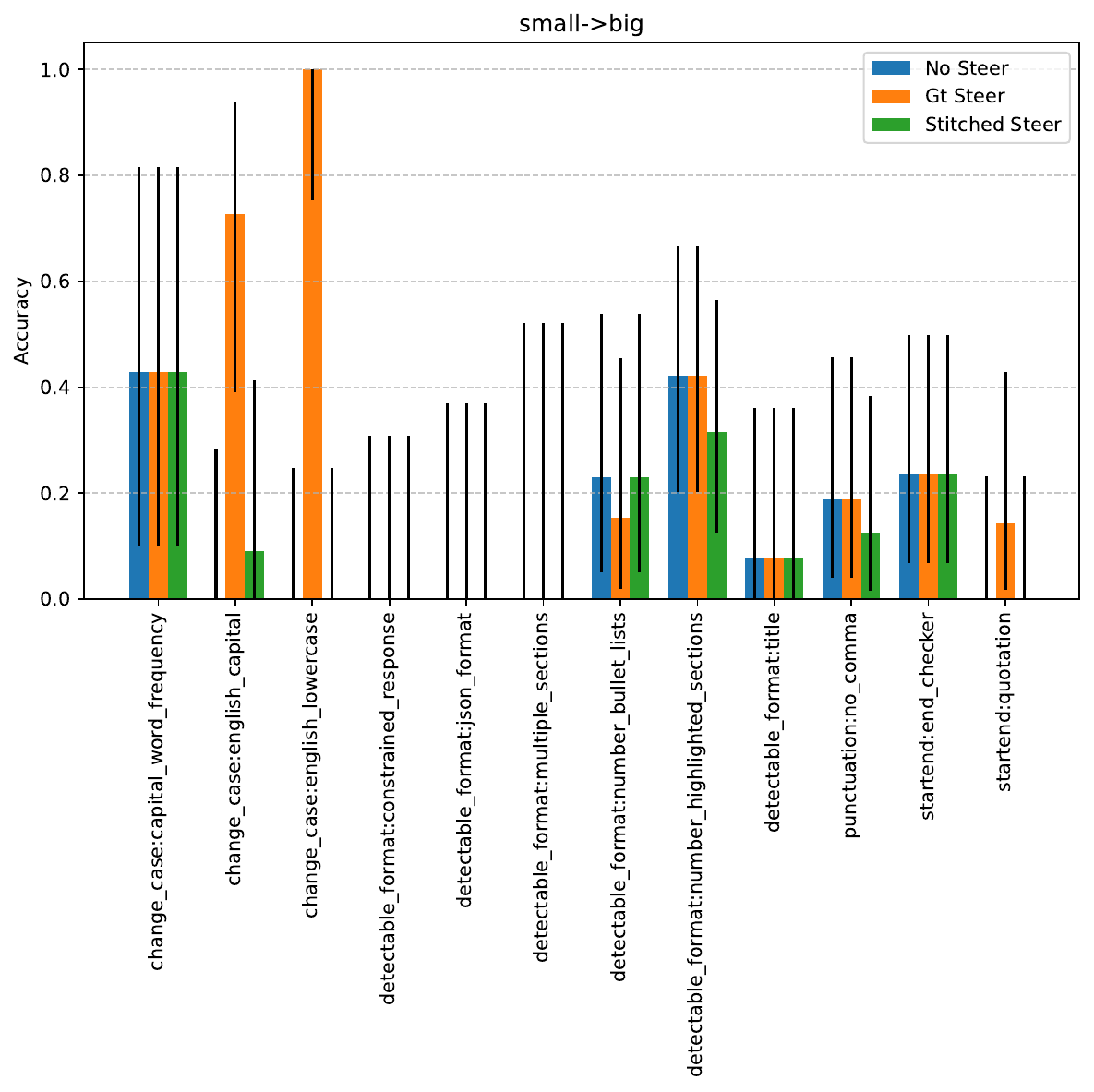}\\
    \includegraphics[width=0.5\linewidth]{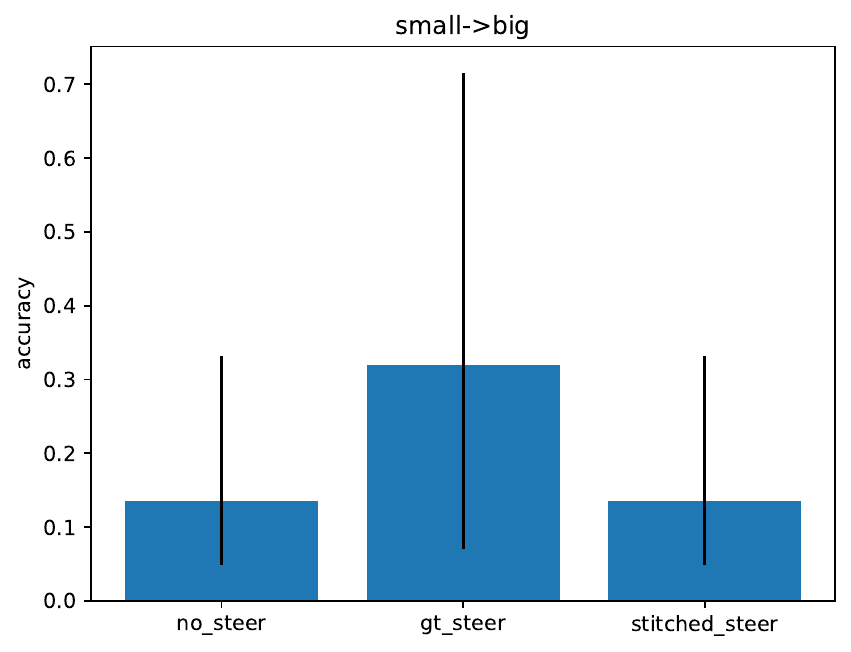}
    \caption{Small to big IFEval steering. Despite strong performance in ground truth steering in many cases, the transferred vectors are unable to identify the correct subspaces to induce the instruction following.}
    \label{fig:ifeval_small_big}
\end{figure}

Our current conjecture for why these results are much weaker than the response language results is that the steering vector is computed on the final token of the input prompt. Previous tokens in the input prompt include chat templating tokens like \texttt{<[start,end]\_of\_turn>}. The stitch is not trained on these tokens (they do not naturally occur in OpenWebText for example) and thus does not learn to properly reconstruct the important components of these tokens, which likely have very complex underlying latent representations but are important for steering. Furthermore, as is clear from the plots, the IFEval dataset is incredibly small so the confidence intervals are wide. The steering vectors are probably noisy for the instructions that have few examples (about 10 pairs) and the noise is compounded through transferring.

\section{Attribution Correlation Histograms}\label{sec:attribution_correlation_histograms}
\begin{figure}[]
    \centering
    \begin{subfigure}[b]{0.32\linewidth}
    \includegraphics[width=\linewidth]{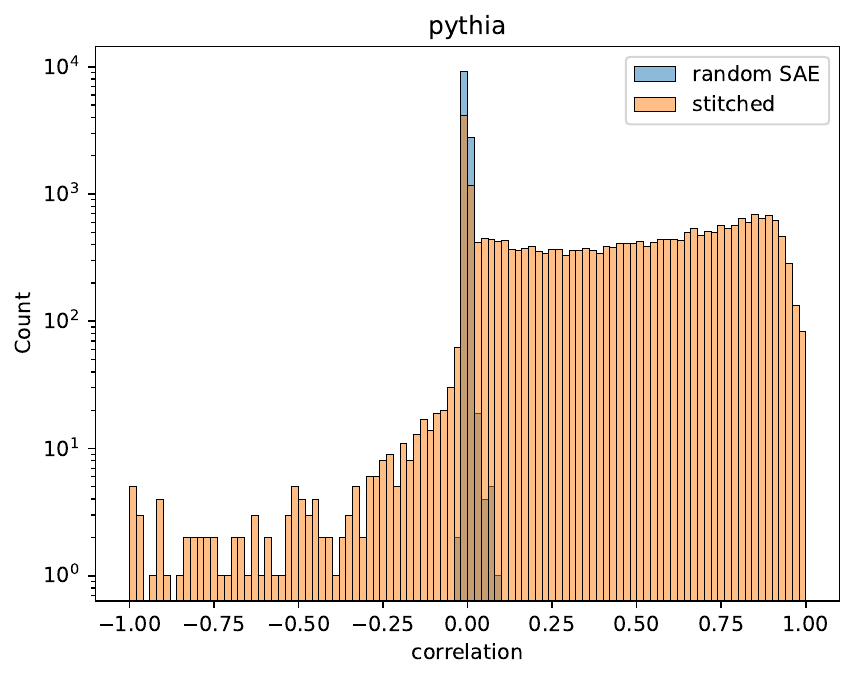}
    \caption{pythia-70m-deduped.3 to pythia-160m-deduped.4}
    \end{subfigure}
    \hfill
    \begin{subfigure}[b]{0.32\linewidth}
    \includegraphics[width=\linewidth]{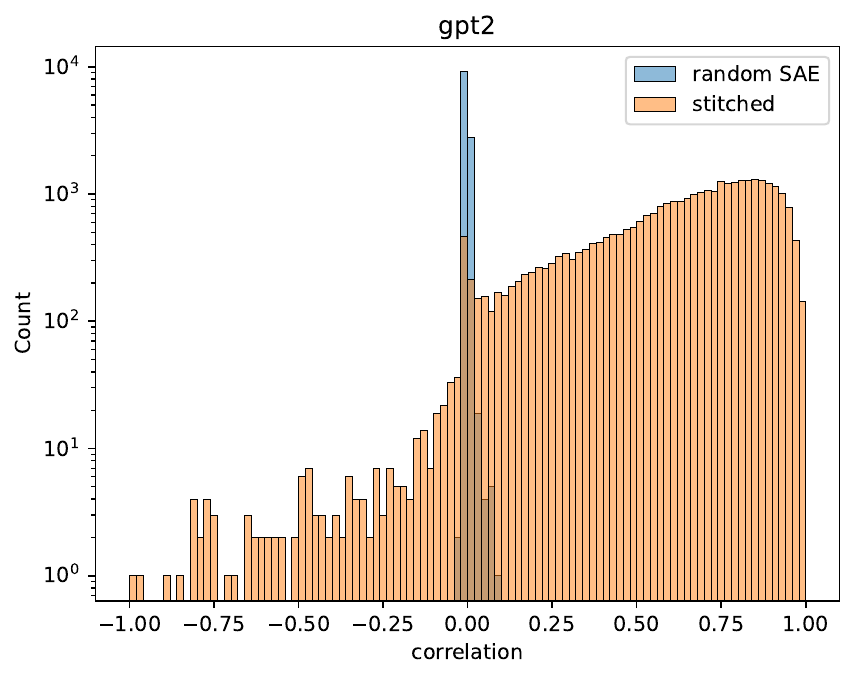}
    \caption{gpt2-small.6 to gpt2-medium.10}
    \end{subfigure}
    \hfill
    \begin{subfigure}[b]{0.32\linewidth}
    \includegraphics[width=\linewidth]{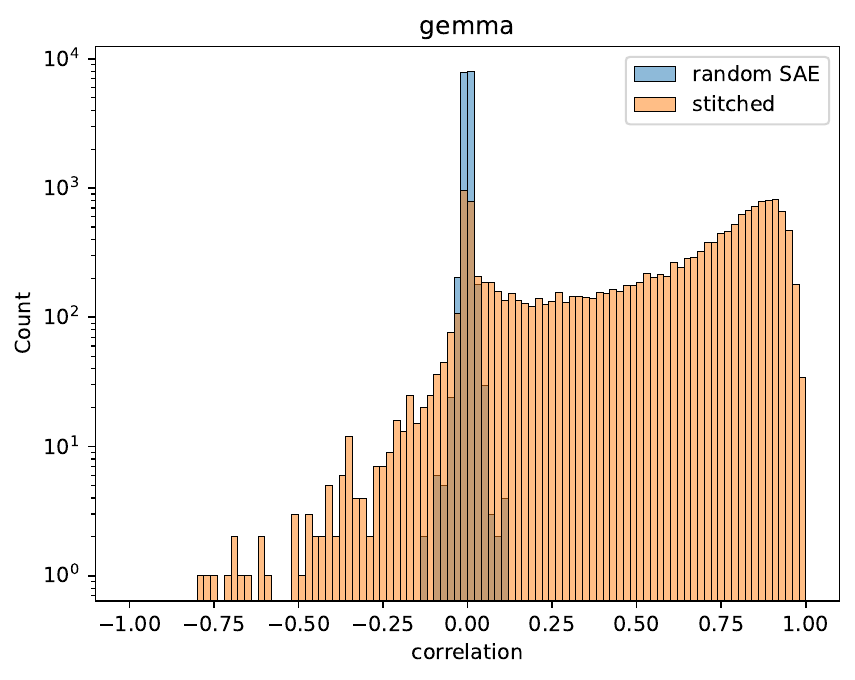}
    \caption{gemma-2-2b.20 to gemma-2-9b.33}
    \end{subfigure}
    \caption{Attribution correlation histograms for three model stitches. We can see that for most features we do much better than a random stitch. However, there are some straggling features that appear to get either flipped or do not transfer well.}
    \label{fig:attribution_correlation_histograms}
\end{figure}

\begin{figure}
    \centering
    \includegraphics[width=0.45\linewidth]{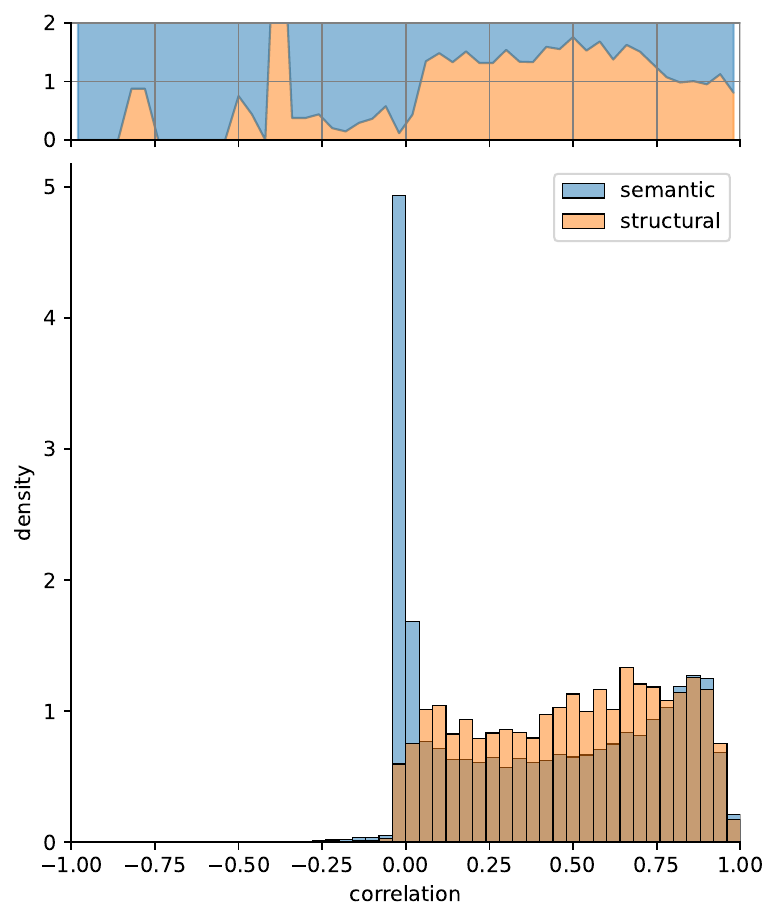}
    \hfill
    \includegraphics[width=0.45\linewidth]{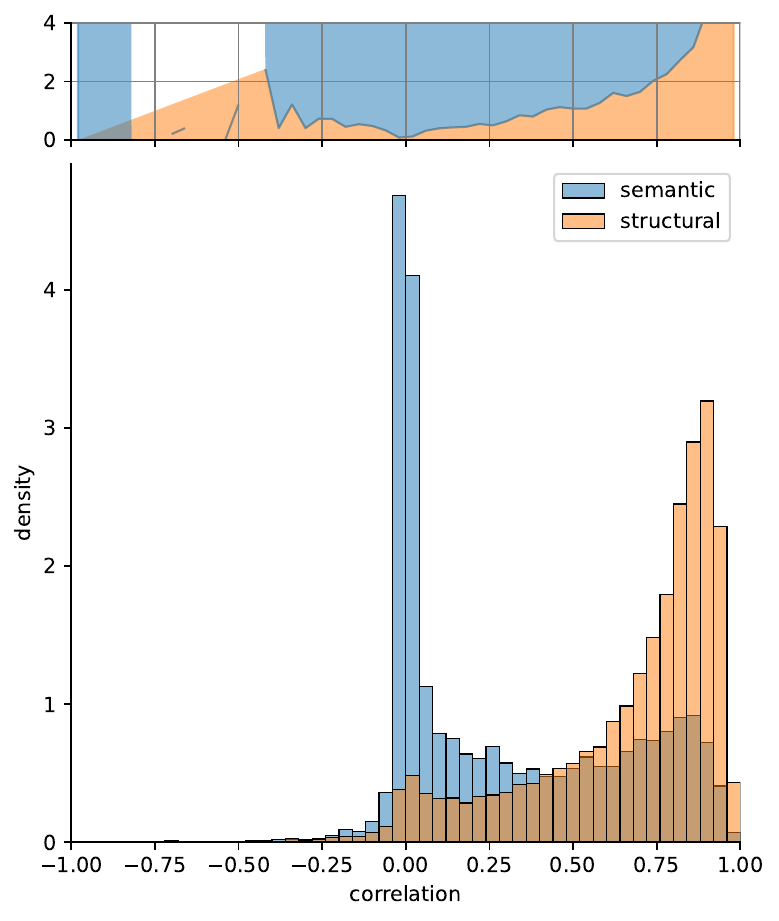}
    \caption{(Left) Pythia and (right) Gemma transfer differences between semantic and structural. We find somewhat similar patterns to the GPT2 case, namely the transferability of structural features and cluster of semantic features that do not transfer appears consistent. Since structural features naturally tend to be higher density, the stitch is incentivized via the MSE penalty to preserve those directions of space because there is higher variance in these dimensions and more consistently help reduce MSE.}
    \label{fig:semantic_structural_pythia_gemma}
\end{figure}

We plot histograms of the attribution correlation metric for our model-layer pairs in Figure \ref{fig:attribution_correlation_histograms}. We plot the semantic/structural difference for Pythia and Gemma in Figure \ref{fig:semantic_structural_pythia_gemma}.

%



\section{Semantic / Structural Augmentation}
\subsection{Prompt}\label{sec:semantic_vs_structural_prompt}
We prompt $\texttt{gemma-2-9b-it}$ $5$ times with temperature $1.0$ using the following prompt (generated using GPT-4 and mildly edited), resulting in $6$ versions of the same sentence. We use context size $128$.
\begin{lstlisting}
Transform the given sentence so that its meaning is completely unrelated, but the syntax, punctuation, and grammatical structure remain identical. This means:
- Keep all function words (e.g., "the," "and," "while") and punctuation (e.g. commas, periods, brackets, parentheses, dashes) unchanged.
- Replace content words (nouns, verbs, adjectives, adverbs) with words from entirely different semantic domains (e.g., "run" -> "melt", "dog" -> "radio").
- Ensure that the sentence remains grammatically correct, even though its meaning is completely different.
- Respond with the new sentence and nothing else.

Examples:
    Input: "The scientist carefully examined the ancient manuscript in the dimly lit library."
    Output: "The firefighter accidentally kicked the broken telephone in the noisy bus station."

    Input: "After the storm passed, the children ran outside to play in the puddles."
    Output: "Once the debate ended, the accountants flew abroad to invest in the factories."

Again, your instructions are:
- Ensure that the core meaning is entirely different from the original sentence.
- Do not use synonyms or words from the same semantic category.
- Maintain identical syntax, punctuation, and grammatical structure.
- The new sentence must be valid and natural, despite the unrelated meaning.
- Respond with the new sentence and nothing else.

Here is the original sentence.
Input: {sentence}
\end{lstlisting}
\subsection{Example Ablations and Classifications}
See Table \ref{tab:example_ablations} for example generated augmentations and Table \ref{tab:cherrypicked_structural_vs_semantic} for examples of classified features and their top activations.
\begin{table}[]
\centering
\caption{Examples of augmented sentences generated by our LLM procedure. Clearly there is some noise (a feature that only activates on ``But" will be classified as structural) but we found these hard to avoid.}
\label{tab:example_ablations}
\begin{tabular}{@{}cl@{}}
\toprule
Augmentation & Sentences                                                                                     \\ \midrule
1            & "But he added that Arizona’s system also created problematic asymmetries and anomalies."      \\
2            & "But she sculpted that broccoli's texture also caused colorful duplicates and polygons."      \\
3            & "But she blended that broccoli also developed confusing melodies and constellations."         \\
4            & "But she designed that polka’s rhythm also fabricated paradoxical similarities and galaxies." \\
5            & "But he measured that broccoli’s shape also produced complicated rhythms and symphonies."     \\
6            & "But she removed that penguin’s melody also produced awkward polygons and tangents."          \\ \midrule
1            & "Use promo code DOUG and play a real money game for FREE!"                  \\
2            & "Bake discount coupon ZEBRA and practice a virtual cooking class for GOLD!" \\
3            & "Order discount coupon ZETA and bake a frozen dinner for DINNER!"           \\
4            & "Use discount code ORANGE and bake a metal game for SILVER!"                \\
5            & "Apply offer code ZEBRA and bake a delicious casserole for DINNER!"         \\
6            & "Activate offer code ZEBRA and purchase a tropical fruit for GRATUITOUS!"   \\ \bottomrule
\end{tabular}
\end{table}
\begin{table}[]
    \centering
    \caption{Examples of structural vs. semantic features based on feature centric evaluations. Drawn from \texttt{pythia-70m-deduped} pre-layer $3$ SAEs. Tokens are highlighted based on activation strength.}
    \resizebox{\textwidth}{!}{\begin{tabular}{llll}
    \toprule
         \textbf{Feature Index} & \textbf{Category} & \textbf{Description} & \textbf{Selected Top Activating Examples} \\
         \midrule
         $18139$ & Semantic & Geometric concepts & \begin{tabular}{@{}c@{}}not passing through any \colorword{84.8}{vertex} of a 
         \colorword{100.0}{triangle}\\the concept of \colorword{62.7}{order} into \colorword{100}{plane} \colorword{59}{geometry}
         \end{tabular} \\
         \midrule
         $4469$ & Semantic & "-bility" words & \begin{tabular}{@{}c@{}} increase brand awareness and desir\colorword{100}{ability} ahead \\
         negative or every obstacle and imposs\colorword{100}{ibility}
         \end{tabular}\\
         \midrule
        $889$ & Structural & Nouns after "the" & \begin{tabular}{@{}c@{}}of the \colorword{71}{confidentiality} relating to the \colorword{100}{substance}-abuse\\the \colorword{100}{plot} beats don’t stray far from the \colorword{82}{genre}\end{tabular}\\
         \midrule
         $11062$ & Structural & Token after "and" list & 
         \begin{tabular}{@{}c@{}} Java, PHP, Python and Ruby \colorword{100}{still} ensconced\\solar, geothermal, hydroelectric, and biomass \colorword{100}{that}\end{tabular}\\
         \bottomrule
    \end{tabular}}
    \label{tab:cherrypicked_structural_vs_semantic}
\end{table}
\section{Attention Patterns}
\begin{figure}[h]
    \centering
    \includegraphics[width=0.7\linewidth]{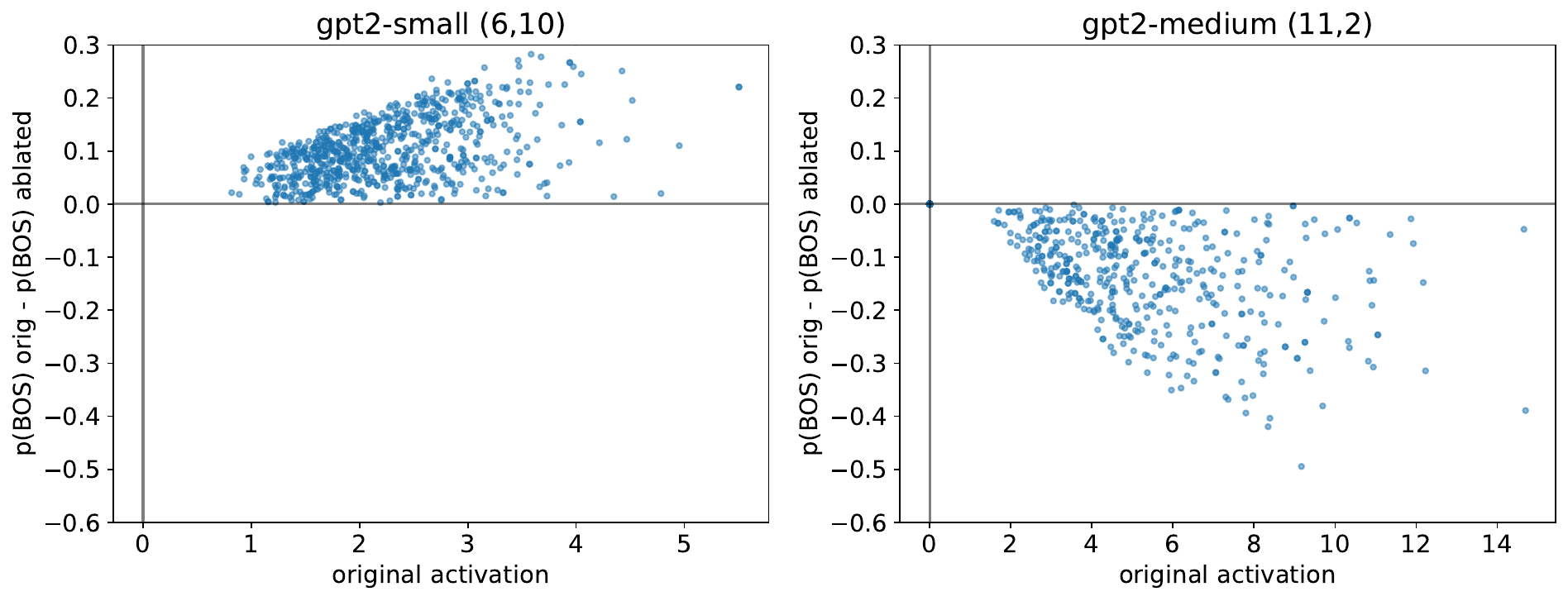}
    \caption{Example of how the same attention deactivation feature from before can act as an activator for a different downstream attention head after transfer.}
    \label{fig:attention_deactivation_flipped}
\end{figure}

See Figure \ref{fig:attention_deactivation_flipped} for an example of the attention deactivation acting as an attention activation feature for an attention head after being transferred.
\section{Compute Statement}
All experiments were run on single Quadro $6000$ / RTX$3090$ (both have 24GB VRAM) configurations. We estimate total GPU hours to be around 10000 in total for the project and most components of the experiments run contiguously for at most $1$ day.

\end{document}